\documentclass[10pt,journal,compsoc]{IEEEtran}
\ifCLASSOPTIONcompsoc
  \usepackage[nocompress]{cite}
\else
  \usepackage{cite}
\fi
\ifCLASSINFOpdf
\else
\fi
\usepackage{microtype}
\usepackage{graphicx}
\usepackage{caption}
\usepackage{nicefrac}
\usepackage{subcaption}
\usepackage{booktabs} 
\usepackage{amsthm, amssymb}
\usepackage{amsmath}
\usepackage{adjustbox}
\usepackage{enumitem}
\usepackage{multirow}
\usepackage{pifont}
\usepackage{color, colortbl}
\usepackage[dvipsnames]{xcolor}
\usepackage[ruled,vlined]{algorithm2e} 

\newtheorem{theorem}{Theorem}[section]

\newtheorem{lemma}[theorem]{Lemma}

\usepackage{hyperref}
\hypersetup{colorlinks, breaklinks, citecolor=[rgb]{0., 0., 0.},
anchorcolor=[rgb]{0.,0.,0.}, linkcolor=[rgb]{0.0, 0., 0.},
urlcolor=[rgb]{0., 0., 0.} 
}

\def\equationautorefname~#1\null{Eq.~(#1)\null}
\newcommand{\aref}[1]{\hyperref[#1]{Appendix~\ref{#1}}}

\newcommand{\ie}{\textit{i}.\textit{e}., }
\newcommand{\eg}{\textit{e}.\textit{g}., }

\begin{document}
%
\title{Converting Artificial Neural Networks to Spiking Neural Networks via Parameter Calibration}
%
%
%
%

\author{Yuhang Li,~\IEEEmembership{Student Member,~IEEE}, Shikuang Deng, Xin Dong, and~Shi~Gu,~\IEEEmembership{Member,~IEEE}.
\IEEEcompsocitemizethanks{\IEEEcompsocthanksitem Y. Li, S. Deng, and S. Gu are with University of Electronic Science and Technology of China, Chengdu, Sichuan 611731, China; Y. Li is also with Yale University, New Haven, CT, 06511.
E-mail: liyuhang699@gmail.com, dengsk119@std.uestc.edu.cn, gus@uestc.edu.cn
\IEEEcompsocthanksitem X. Dong is with Harvard University, Cambridge, MA, 02138. 
E-mail: xindong@g.hardvard.edu}

\thanks{Manuscript received April 19, 2005; revised August 26, 2015.}}

%
%

\markboth{Preprint, work in progess. }%
{Shell \MakeLowercase{\textit{et al.}}: Bare Demo of IEEEtran.cls for Computer Society Journals}
%



\IEEEtitleabstractindextext{%
\begin{abstract}
Spiking Neural Network (SNN), originating from the neural behavior in biology, has been recognized as one of the next- generation neural networks. Conventionally, SNNs can be obtained by converting from pre-trained Artificial Neural Networks~(ANNs) by replacing the non-linear activation with spiking neurons without changing the parameters. In this work, we argue that simply copying and pasting the weights of ANN to SNN inevitably results in activation mismatch, especially for ANNs that are trained with batch normalization~(BN) layers. To tackle the activation mismatch issue, we first provide a theoretical analysis by decomposing local conversion error to clipping error and flooring error, and then quantitatively measure how this error propagates throughout the layers using the second-order analysis. Motivated by the theoretical results, we propose a set of layer-wise parameter calibration algorithms, which adjusts the parameters to minimize the activation mismatch. Extensive experiments for the proposed algorithms are performed on modern architectures and large-scale tasks including ImageNet classification and MS COCO detection. We demonstrate that our method can handle the SNN conversion with batch normalization layers and effectively preserve the high accuracy even in 32 time steps. For example, our calibration algorithms can increase up to 65\% accuracy when converting VGG-16 with BN layers.

\end{abstract}

}
\maketitle

\IEEEdisplaynontitleabstractindextext

%
\IEEEpeerreviewmaketitle

\IEEEraisesectionheading{\section{Introduction}\label{sec:introduction}}

%
%
%
%
\IEEEPARstart{H}{uman} 
brain is recognized as an extremely astounding intelligent system that regulates its connection across multiple neuronal layers in response to target tasks.
Inspired by this, Artificial Neural Networks (ANNs) emulate the structural characteristic of the visual cortex, achieving beyond-human-level performance on various tasks \cite{silver2016mastering,he2016deep,vinyals2019grandmaster}.
However, the great success of ANNs is highly based on intensive computation~\cite{sze2017efficient}, while a human brain requires only a small amount of power budget ($\sim$20 Watts)~\cite{cox2014neural} and performs asynchronous computation. Exploring next-generation brain-inspired intelligent models will not only benefit to real-world applications considering the potential resource-saving but also advance the research of artificial intelligence.

In this context, Spiking Neural networks~(SNNs) have gained increasing attention for building low-power and biomimicry intelligence \cite{christensen20222022,roy2019scaling,DENG2020294,roy2019nature}.
Different from ANNs, SNNs consist of biological   neurons~\cite{hodgkin1952quantitative,izhikevich2003simple} that convey information through the time domain.
At each time step, each neuron generates a spike if the accumulated membrane potential exceeds a firing threshold, otherwise, it would stay inactive. 
Such spiking neuronal behavior enables binarized activation in networks, thus bringing an advantage of energy efficiency in SNNs. 
Existing studies have shown that the event-driven behavior of SNNs can be implemented on emerging neuromorphic hardware to yield orders of magnitude energy savings over ANNs \cite{akopyan2015truenorth,davies2018loihi,furber2014spinnaker,pei2019towards}.
Besides energy efficiency, SNNs also have an inherent ability to process spatial-temporal information. 
By adjusting the time steps to process in SNNs, one can achieve flexible trade-offs between task performance and inference latency. 

Despite its energy-efficiency and special spatial-temporal processing ability, training SNNs is challenging because that spiking neurons fire discrete spikes whose gradients are not well-defined. As a result, applying gradient-based optimization for SNNs could be difficult and unstable.
Various training methods have been proposed to address the training problems.
For example, spike-timing-dependent plasticity (STDP)  \cite{bi1998synaptic} approaches \cite{lee2018training,iakymchuk2015simplified,lobov2020spatial} follow a bio-plausible unsupervised Hebbian learning rule \cite{hebb2005organization} and update weights without gradients. However, STDP methods are limited to small-scale datasets (\ie MNIST) because of the absence of global optimization rule.
Another representative type of training methods is to use surrogate gradient function for spiking neurons.
These methods design differentiable surrogate activation functions (and corresponding surrogate gradients) to optimize SNNs~\cite{lee2016training,lee2020enabling,neftci2019surrogate,shrestha2018slayer,gu2019stca,wu2018spatio,li2021differentiable}.
They provide good performance with small number of timesteps (\ie 5 $\sim$ 20) even on the large-scale dataset such as ImageNet~\cite{deng2009imagenet}. 
However, they require tremendous computational hardware resources for training because of large computational graph from multi-timestep operations.
In a word, training SNNs from scratch remains challenging and expensive.

\begin{figure*}
    \centering
    \includegraphics[width=0.85\textwidth]{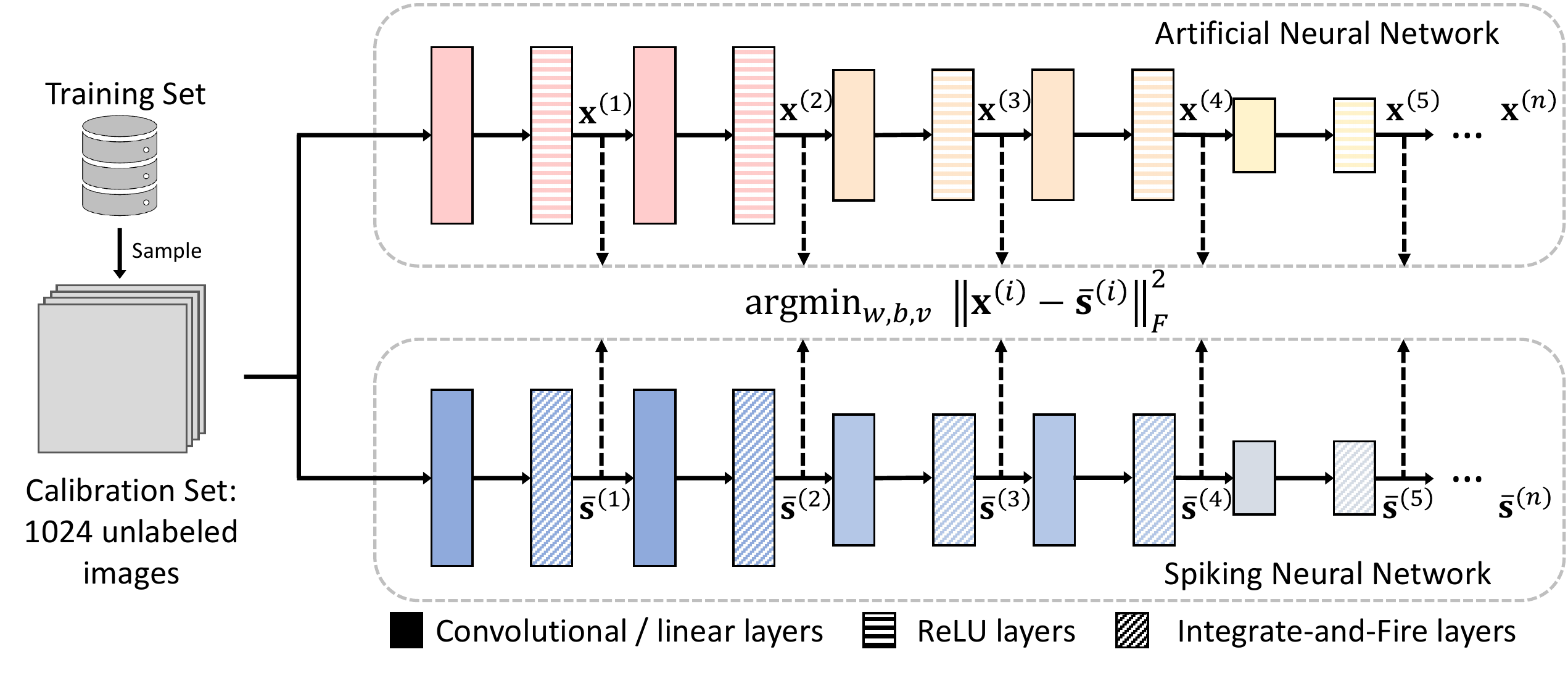}
    \caption{An overview of our proposed algorithm. Using a small subset of training data (1024 unlabeled images), we calibrate the parameters, including weights, bias, initial membrane potentials, to match the activation between ANNs and SNNs. }
    \label{fig_intro}
\end{figure*}

Besides directly training SNNs from scratch, obtaining a high-performance SNN by converting a pretrained ANN~\cite{diehl2016conversion,rueckauer2017conversion,Sengupta2018Going} is another promising direction with several advantages. It circumvents the gradient estimation problem because it adapts the well-established ANN learned features, leading to high scalability in complex tasks. 
To perform the conversion, prior approaches convert the ReLU activation in pre-trained ANN to the spiking neuron in SNNs with methods like threshold balancing~\cite{Diehl2015Fast,Sengupta2018Going}, weight normalization~\cite{rueckauer2017conversion}, and soft-reset mechanism~\cite{rueckauer2017conversion, han2020rmp}.
However, nearly all of them fail to convert an ANN into a SNN within extremely low time steps, especially when the ANN contains batch normalization (BN) layers, (\eg RMP~\cite{han2020rmp} obtains only $\nicefrac{2}{3}$ accuracy of ANN with 32 time steps when converting VGG-16 on CIFAR-10).

We argue that the major bottleneck for ANN-SNN conversion in extremely low time steps is the \textbf{activation mismatch} issue. Although previous efforts study how to approximate ANN using SNN, they still copy-and-paste parameters from ANN to SNN, simply considering the minimization of conversion error in parameter space. This minimization cannot ensure the statistical distance characterized by the output distribution on a group of output nodes since the activation functions in the two cases are different in nature. As a result, the ANN and converted SNN do not share similar network output. 
To empirically study the activation mismatch issue, we convert a VGG-16 to SNN with 16 time steps using the conventional copy-paste methods. Then, we measure the activation distance (detailed description is in \autoref{sec_problems}) between the original VGG-16 and spiking VGG-16 in each layer. In \autoref{fig_relative_err}, we observe that activation mismatch steadily increases, and the mismatch is much higher if the ANN has BN layers. Unfortunately, BN layers are believed to be crucial for ANNs to achieve high accuracy~\cite{santurkar2018does}. Therefore, it is compelling to study how to convert ANN with BN layers in extremely low time steps.

In order to theoretically characterize the conversion error between SNN and ANN, we first formulate the conversion error brought by the integrate and fire (IF) neurons in SNNs. Then we analyze how each layer's error propagates throughout the network using the second-order analysis~\cite{dong2017learning}, showing that the global output error could be bounded by the weighted sum of the local conversion error. 
This motivates us to come up with a set of parameter calibration algorithms.
As shown in \autoref{fig_intro}, starting from the first layer to the last layer, the parameter calibration fine-tunes the parameters, including weights, bias, and membrane potentials, to minimize the local conversion error. 
Unlike previous conversion work which simply \textbf{transplants the weights parameters} to the SNN, in this work we show that the \textbf{activation transplant} is much more important.
Moreover, the calibration algorithm is fast and simple to use, without any modifications in the source ANN pre-training and requires only 1024 unlabeled training images. Calibrating the bias only takes several minutes even on a CPU. 

The novel contributions of the paper are threefold:

\begin{itemize}[leftmargin=*]
    \item We formulate the conversion equation and divide the conventional conversion error into flooring and clipping errors. Then we analyze the error propagation through layers using the second-order gradient information (Hessian matrix). Results show that conversion error can be bounded by the weighted sum of local conversion error in each layer. 
    \item We propose a layer-wise calibration algorithm to adjust the network parameters including weights, bias, and initial potential to diminish the conversion error. To accommodate different user requirements, we provide Light Pipeline and Advanced Pipeline to balance accuracy and practical utility.
    \item We verify our algorithms on large-scale datasets like ImageNet~\cite{deng2009imagenet}. In addition to ResNets and VGG networks in previous work, we test a lightweight model MobileNet~\cite{Howard2017MobileNetsEC}, a NAS searched model MNasNet~\cite{tan2019mnasnet}, and a large model RegNetX-4GF~\cite{radosavovic2020regnet} (80.0\% top-1 accuracy) for the first time in {ANN-to-SNN} conversion. We also include the experiments in COCO object detection tasks with RetinaNet~\cite{lin2017focal} and Faster R-CNN~\cite{ren2015faster}.
\end{itemize}

\section{Related Work}

\subsection{Direct Training SNN}
For training-based SNN, there are supervised and unsupervised learning paradigms. Unsupervised learning is based on synaptic plasticity, which is related to the learning scheme in real biological behavior. Based on time sensitivity and the two neurons' firing time interval, the synaptic plasticity can either strengthen or weaken the connection weight~\cite{kheradpisheh2018stdp,iyer2020classifying,li2019research}. However, these methods are still limited to small-scale tasks, such as the MNIST dataset. 
On the other hand, supervised learning seeks to use gradient descent~\cite{tavanaei2019deep} like ANN. Surrogate gradient (spiking-based backpropagation) methods use a soft relaxed function to replace the hard step function and train SNN like RNN~\cite{wu2018spatio,shrestha2018slayer}. They suffer from expensive computation during the training process on complex network architecture\cite{rathi2019enabling}. 
Choosing the appropriate surrogate gradient function is also extensively studied: \cite{li2021differentiable} optimizes the surrogate gradient using the finite difference method, \cite{wu2022brain} uses meta learning to estimate the function of spiking neurons.  
PLIF~\cite{fang2021incorporating} proposes to learn the parameters in LIF neurons via gradient descent. TdBN, BNTT~\cite{zheng2020going, kim2020revisiting} improve direct training by adapting batch normalization layers in SNNs. 
TET~\cite{deng2022temporal} studies the choice of the loss function to provide better convergence in SNNs. \cite{kim2022neural} employs neural architecture search tailored for SNN. 

\subsection{ANN-SNN Conversion}

Due to the time dimension of SNN, the direct training of an SNN usually requires even more GPU hours than training an ANN from scratch. 
ANN-to-SNN conversion methods, thus, have become the most efficient way to obtain a well-performed SNN. Methods like data-based normalization~\cite{Diehl2015Fast,rueckauer2016theory} or threshold balancing~\cite{Diehl2015Fast,diehl2016conversion} study how to determine the spiking neurons configurations, \eg number of time steps, threshold, and leaky factor. 
The major bottleneck of these prior arts is the costly inference latency required to preserve high accuracy. As an example, \cite{Sengupta2018Going} requires more than 2k time steps to get results as accurate as of the source ANN. 
Recently, several methods have been proposed to reduce the conversion loss and number of time steps. The soft-reset (\textit{a.k.a.}, the reset-by-subtraction mechanism) is the most common technique to address the potential reset's information loss \cite{rueckauer2016theory,han2020deep}. Our IF neuron model also adopts this strategy. \cite{rueckauer2017conversion} suggests using a percentile threshold, which avoids picking the outlier in the activation distribution.
Spike-Norm~\cite{Sengupta2018Going} tests deep architectures like VGG-16 and ResNet-20. In this work, we further extend the source architecture to more advanced architectures like MobileNet, RegNet, and MNasNet.
RMP~\cite{han2020rmp} and TSC~\cite{han2020deep} achieve near-to-origin accuracy by adjusting the threshold according to the input and output spike frequency.
\cite{deng2021optimal, li2021free} decompose the conversion loss into each layer and reduce it via shifting bias. Low latency converted SNN is an ongoing research challenge since it still requires a considerable amount of simulation length. 
We argue that the minimization in weight space does not equal the minimization in network output space, therefore, it is necessary to calibrate the parameters to match ANN and SNN. 

Lately, there is a trend for ``hybrid'' learning, \ie leveraging both ANN's feature and gradient descent to obtain an SNN. The Hybrid Training~\cite{rathi2020enabling} proposes a two-stage method: first converting an ANN to an SNN, and then fine-tuning this SNN using the surrogate gradient methods.
Spiking Tandem Learning~\cite{wu2019tandem, wu2021progressive} adopts a progressive approach to learn the SNN layer-by-layer.
It can be viewed as a type of knowledge transfer\cite{romero2014fitnets}. 
Our method also builds the spiking neural network layer-by-layer. However, we should emphasize that our method is motivated by a rigorous theoretical analysis of conversion. Additionally, we do not employ end-to-end training pipelines on SNNs, and our method only requires a tiny subset from the training dataset (typically 1024 unlabeled images) and can be finished in several minutes.

\subsection{Second-order Analysis}

Using Taylor series expansion, especially with the second-order term, is useful for analyzing the curvature of a computational model.
The history of second-order information in perturbation analysis can be traced to the 1990s like Optimal Brain Surgeon~\cite{hassibi1993optimalbrain, dong2017obslayer}. 
The Hessian matrix is essential for applications that need to impose noise/perturbation to the parameters. For example, in pruning and quantization, the weights parameters need to be removed or rounded to integers. In those cases, the Hessian matrix provides a quantitative measure of loss change after the perturbation. The Fisher pruning~\cite{theis2018faster} selects the weights that have the lowest importance (measured by fisher information) to prune, thus having a minimal effect on the whole network. Hessian-aware weight quantization~\cite{dong2019hawq} computes the largest eigenvalue of Hessian to determine the sensitivity and then determines the suitable bit-width.
BRECQ~\cite{li2021brecq} approximates the second-order error by decomposing it to each block of the neural network. 
In this work, we also make use of second-order analysis to understand the conversion error in activations.

\section{Preliminaries}
\label{sec_pre}
\subsection{Neuron Model for ANN.} 
Consider the $\ell$-th fully-connected layer or convolutional layer in an $n$-layer ANN, we formulate its forwarding process as,
\begin{equation}
    \mathbf{x}^{(\ell+1)} = h(\mathbf{z}^{(\ell)})=h(\mathbf{W}^{(\ell)}\mathbf{x}^{(\ell)}),  1\le\ell\le n, 
    \label{eq_ann}
\end{equation}
where $\mathbf{x}^{(\ell)}$, $\mathbf{W}^{(\ell)}$ denote the input activation and weight parameters in that layer respectively, and $h(\cdot)$ is the ReLU activation function (\ie $\max(0, x)$). One can optionally train a bias parameter $\mathbf{b}^{(\ell)}$ and add it to pre-activation $\mathbf{z}^{(\ell)}$.

\subsection{Neuron Model for SNN.} 
Here we use the Integrate-and-Fire (IF) neuron model \cite{liu2001spike,barbi2003stochastic}. Concretely, suppose at time step $t$ the spiking neurons in layer $\ell$ receive its binary input $\mathbf{s}^{(\ell)}(t)\in \{0, V_{th}^{(\ell-1)}\}$, the neuron will update its temporary membrane potential by,
\begin{equation}
    \mathbf{v}^{(\ell)}_{temp}(t+1) = \mathbf{v}^{(\ell)}(t) + \mathbf{W}^{(\ell)}\mathbf{s}^{(\ell)}(t),
    \label{eq_integrate}
\end{equation}
where $\mathbf{v}^{(\ell)}(t)$ denotes the membrane potential at time step $t$, and $\mathbf{v}^{(\ell)}_{temp}(t+1)$ denotes the intermediate variable that would be used to determine the update from $\mathbf{v}^{(\ell)}(t)$ to $\mathbf{v}^{(\ell)}(t+1)$.
If this temporary potential exceeds a pre-defined threshold $V_{th}^{(\ell)}$, it would produce a spike output $\mathbf{s}^{(\ell+1)}(t)$ with the value of $V_{th}^{(\ell)}$. Otherwise, it would release no spikes, \ie $\mathbf{s}^{(\ell+1)}(t) = 0$. The membrane potential at the next time step $t+1$ would then be updated by \textit{soft-reset} mechanism, also known as \textit{reset-by-subtraction}. Formally, we describe the updating rule as 
\begin{equation}
    \mathbf{v}^{(\ell)}(t+1) = \mathbf{v}^{(\ell)}_{temp}(t+1) - \mathbf{s}^{(\ell+1)}(t), \label{eq_snn_neuron}
\end{equation}
\begin{equation}
    \mathbf{s}^{(\ell+1)}(t) = \begin{cases}
     V_{th}^{(\ell)} & \text{if }\mathbf{v}^{(\ell)}_{temp}(t+1) \ge V_{th}^{(\ell)} \\
    0 & \text{otherwise } 
    \end{cases}.
    \label{eq_if}
\end{equation}
Note that $V_{th}^{\ell}$ can be distinct in each layer. Thus, we cannot represent the spike in the whole network with binary signals. 
This problem can be avoided by utilizing a weight normalization technique to convert the $\{0, V_{th}^{(\ell-1)}\}$ spike to $\{0,1\}$ spike in every layers, given by:
\begin{equation}
    \mathbf{W}^{(\ell)} := \frac{V_{th}^{(\ell-1)}}{V_{th}^{(\ell)}}\mathbf{W}^{(\ell)}, \ \ \ V_{th}^{(\ell)}:= 1. 
\end{equation}
Recursively applying the above equalization, we can use 0,1 spike to represent the intermediate activation for each layer. For the rest of the paper, we shall continue using the notation of $\{0, V_{th}^{(\ell-1)}\}$ spike for less ambiguity.

As for the input to the first layer and the output of the last layer, we do not employ any spiking mechanism. We use the first layer to direct encode the static image to temporal dynamic spikes, this can prevent the undesired information loss of the encoding. For the last layer output, we only integrate the pre-synaptic input and do not fire any spikes. This is because the output can be either positive or negative, yet \autoref{eq_if} can only convert the ReLU activation. 

\subsection{Converting ANN to SNN. }
Compared with ANN, SNN employs binary activation (\ie spikes) at each layer. To compensate for the loss in representation capacity, researchers introduce the time dimension to SNN by repeating the forwarding pass $T$ times to get final results. Ideally, to convert the ANN to an SNN, the conversion is expected to have approximately the same input-output function mapping as the original ANN in the final output, \ie,
\begin{equation}
    \mathbf{x}^{(n)} \approx \bar{\mathbf{s}}^{(n)}= \frac{1}{T}\sum_{t=0}^T\mathbf{s}^{(n)}(t).
    \label{approx_output}
\end{equation}
In practice, the above approximation only holds when $T$ grows to 1k or even higher. However, high $T$ would lead to large inference latency thus diminishing SNN's practical utility.

\subsection{Existing Problems of ANN-SNN Conversions.}
\label{sec_problems}

\begin{figure}[t]
    \centering
    \includegraphics[width=\linewidth]{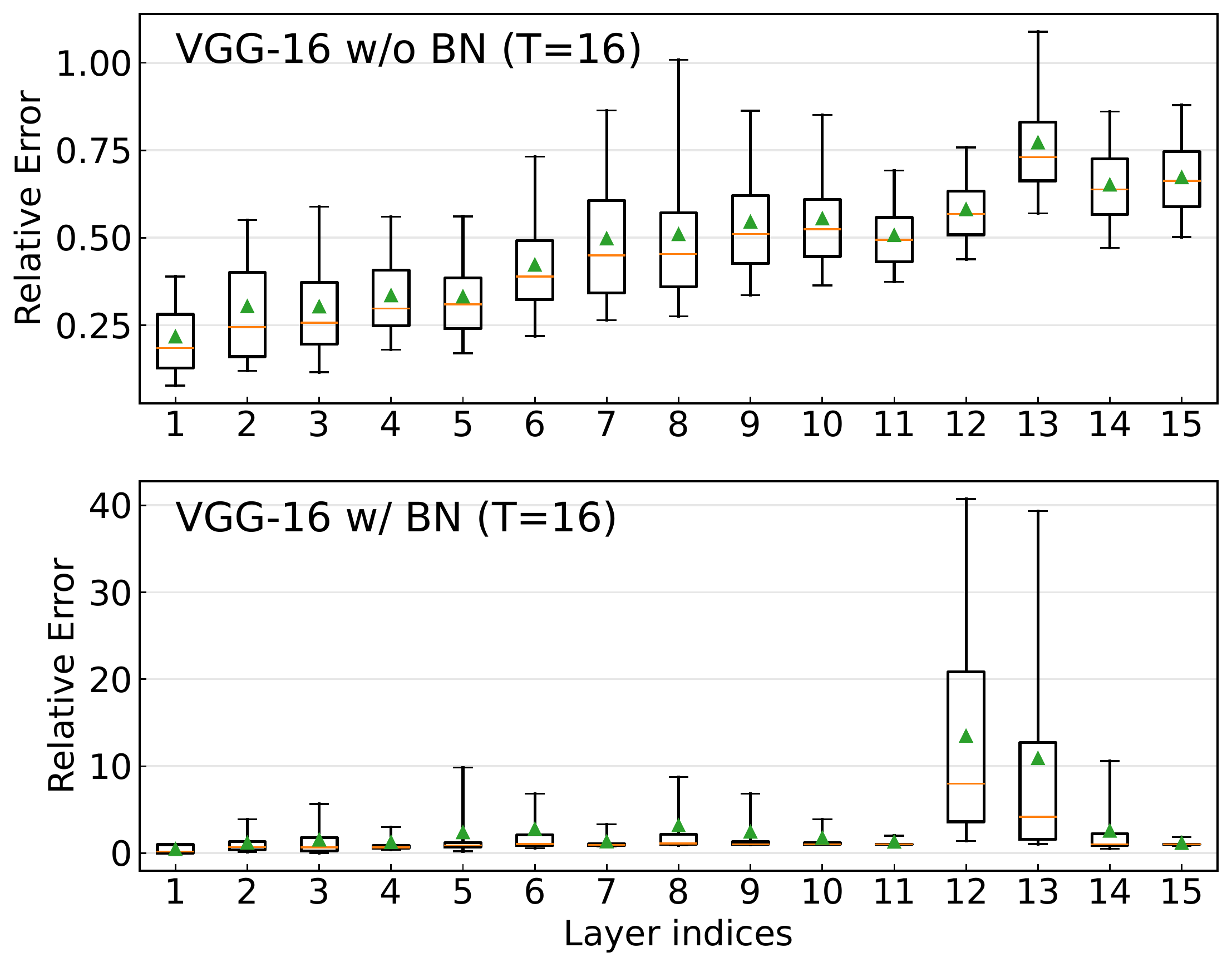}
    \caption{We measure the relative error using the conventional copy-paste conversion. In each layer, we calculate the relative error in \emph{every channel} and put them in a boxplot. The whisker shows the range between 5-th and 95-th percentile numbers. The orange line ``\textcolor{orange}{$-$}" denotes the median number and the green triangle ``{\color{ForestGreen}{$\blacktriangle$}}" denotes the mean value. }
    \label{fig_relative_err}
\end{figure}

Existing methods for converting ANNs to SNNs can be summarized in two steps. First, the firing threshold ($V_{th}$) in each IF neuron layer is determined. And second, the ANN parameters are simply copied-pasted to SNN.
Unfortunately, these methods usually fail to convert the ANN to SNN in low time steps $(T\le 128)$. We argue that the major problem is the \emph{activation mismatch}, that is, even if we manage to obtain the optimal firing threshold, the converted SNN still discretizes the activation in limited time steps. This discretization changes the activation distribution and has a cumulative effect when the activation propagates through the network.
To quantitatively demonstrate the activation mismatch, we calculate the relative error in each layer, \ie $\frac{||\mathbf{x}-\bar{\mathbf{s}}||_F^2}{||\mathbf{x}||_F^2}$, and visualize it in \autoref{fig_relative_err}. If the VGG-16 is trained without batch normalization~\cite{ioffe2015batchnorm} (BN) layers, we observe that relative error initially is low can gradually increase to 1.
However, when enabling BN layers, the relative error is significantly larger. In the penultimate layer, the relative error even reaches over 10 in some channels, thus causing a substantial accuracy deficit. Nevertheless, BN layers are essential for achieving high performance on large-scale dataset like ImageNet~\cite{deng2009imagenet}. In this paper, we aim to convert ANNs to SNNs \emph{especially with BN layers}. 

\section{Methodology}

In this section, we introduce our ANN-SNN conversion framework in detail. First, \autoref{sec_local_error} analyzes the conversion error between ReLU activation and IF spikes. In \autoref{sec_err_analysis}, we study the conversion error in output space using second-order analysis to provide insights into error reduction.
\autoref{sec_cr_at} presents two simple adjustments to improve ANN-SNN conversion. 
Next, based on our theoretical analysis, we propose the parameter calibration method to minimize the layer-wise objective.

\label{sec_method}
\subsection{Dissecting the Conversion Error}
\label{sec_local_error}

To understand how IF neurons differ from ReLU neurons, we theoretically analyze the conversion error. 
We first derive the relationship between expected spikes in two consecutive layers, \ie $\bar{\mathbf{s}}^{(\ell)}$ and $\bar{\mathbf{s}}^{(\ell+1)}$. 
Suppose the initial membrane potential $\mathbf{v}^{(\ell)}(0) = \mathbf{0}$. Substitute \autoref{eq_integrate} into \autoref{eq_snn_neuron} and sum over $T$, then we get
\begin{equation}
   \mathbf{v}^{(\ell)}(T) = \mathbf{W}^{(\ell)}\left(\sum_{t=0}^{T}\mathbf{s}^{(\ell)}(t) \right) - \sum_{t=0}^{T}\mathbf{s}^{(\ell+1)}(t).
   \label{eq_T_total}
\end{equation}
Since at each time step, the output can be either $0$ or $V_{th}^{(\ell)}$, the accumulated output $\sum_{t=0}^{T}\mathbf{s}^{(\ell+1)}(t)$ can be written to $mV_{th}^{(\ell)}$ where $m \in \{0, 1, \dots, T\}$ denotes the total number of spikes. Here we assume the terminal membrane potential $\mathbf{v}^{(\ell)}(T)$ lies within the range $[0, V_{th}^{(\ell)})$.
Therefore, according to \autoref{eq_T_total}, we have
\begin{equation}
    T\mathbf{W}^{(\ell)} \bar{\mathbf{s}}^{(\ell)} - V_{th}^{(\ell)} < mV_{th}^{(\ell)} \le T\mathbf{W}^{(\ell)}\bar{\mathbf{s}}^{(\ell)}.
\end{equation}
where $\bar{\mathbf{s}}^{(\ell)}$ is defined in \autoref{approx_output}. Then, we can use floor operation and clip operation to determine the the total number of spikes $m$,
\begin{equation}
    m = \mathrm{Clip}\left(\left\lfloor\frac{T}{V_{th}^{(\ell)}}\mathbf{W}^{(\ell)}\bar{\mathbf{s}}^{(\ell)}\right\rfloor, 0, T\right).
\end{equation}
Here the clip function sets the upper bound $T$ and the lower bound $0$. Floor function $\lfloor x \rfloor$ returns the greatest integer that is less than or equal to $x$. Given this formula, we can calculate the expected output spike,
\begin{align}
   \bar{\mathbf{s}}^{(\ell+1)} & = \mathrm{ClipFloor}(\mathbf{W}^{(\ell)}\bar{\mathbf{s}}^{(\ell)}, T, V_{th}^{(\ell)}) \nonumber\\ 
   & = \frac{V_{th}^{(\ell)}}{T}\mathrm{Clip}\left(\left\lfloor\frac{T}{V_{th}^{(\ell)}}\mathbf{W}^{(\ell)}\bar{\mathbf{s}}^{(\ell)}\right\rfloor, 0, T\right)
   \label{eq_clip_floor}
\end{align}
According to \autoref{eq_clip_floor}, the conversion loss (difference between $\mathbf{x}^{(\ell+1)}$ and $\bar{\mathbf{s}}^{(\ell+1)}$) comes from two aspects, namely the \emph{flooring error} and the \emph{clipping error}.

\begin{figure}[t]
    \centering
    \includegraphics[width=0.8\linewidth]{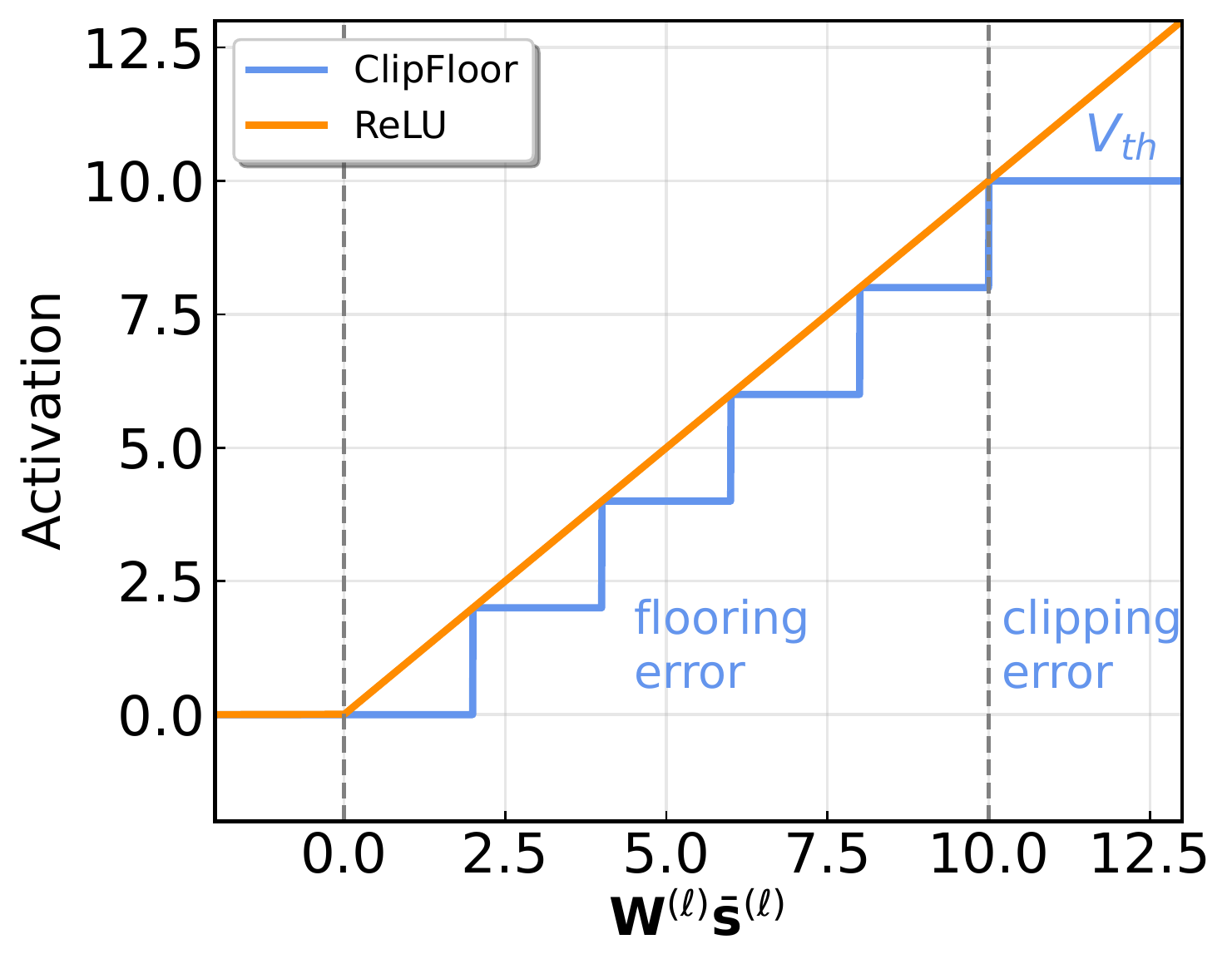}
    \caption{The conversion error between the ReLU activation used in ANN and the output spike in SNN ($V_{th}=10, T=5$) contains flooring error and clipping error.}
    \label{fig_clip_floor}
\end{figure}

In \autoref{fig_clip_floor}, we further indicate that $V_{th}^{(\ell)}$ is crucial for conversion loss because it affects both the flooring and the clipping errors. Increasing $V_{th}^{(\ell)}$ leads to a lower clipping error but a higher flooring error.
Previous work \cite{Diehl2015Fast,diehl2016conversion} sets $V_{th}^{(\ell)}$ to the maximum pre-activations across samples in ANN to eliminate the clipping error. However, the maximum pre-activations are usually outliers. Given this insight, the outliers may tremendously increase the flooring error. As a result, they have to use a very large $T$ (\eg 2000) to decrease the flooring error. 

\subsection{Error Propagation}
\label{sec_err_analysis}
In this section, we extend the previous analysis to the network final output space, \ie $\mathbf{x}^{(n)}$. 
Assuming an ANN is converted to an SNN with its parameters intact, yet all the activation functions are changed from $\mathrm{ReLU}$ to $\mathrm{ClipFloor}$, \ie \autoref{eq_clip_floor}, we are interested in how to measure and minimize the change in the final network output. Let $L(\mathbf{a}) = ||\mathbf{x}^{(n)}-\mathbf{a}||_F^2$ be the loss function of the Euclidean distance between arbitrary vector $\mathbf{a}$ and the original ANN output $\mathbf{x}^{(n)}$, the objective of conversion is to find an optimal SNN average spikes $\bar{\mathbf{s}}^{(n)}$ which minimizes the output discrepancy, 
\begin{equation}
    \min_{\bar{\mathbf{s}}^{(n)}} L(\bar{\mathbf{s}}^{(n)})-L(\mathbf{x}^{(n)}) = \min_{\bar{\mathbf{s}}^{(n)}}||\mathbf{x}^{(n)}-\bar{\mathbf{s}}^{(n)}||_F^2
\end{equation}
We adopt the Taylor series expansion to analyze the conversion error, given by
\begin{equation}
    L(\bar{\mathbf{s}}^{(n)})-L(\mathbf{x}^{(n)}) = \mathbf{e}^{(n),\top}\mathbf{g}^{(n)}+\frac{1}{2}\mathbf{e}^{(n), \top}\mathbf{H}^{(n)}\mathbf{e}^{(n)},
\end{equation}
where $\mathbf{e}^{(n)}=\mathbf{x}^{(n)}-\bar{\mathbf{s}}^{(n)}$ is the difference between ANN activation and (approximated) SNN average spikes. We define $\mathbf{g}^{(n)}=\nabla_{\mathbf{x}^{(n)}}L$ and $\mathbf{H}^{(n)}=\nabla_{\mathbf{x}^{(n)}}^2L$ as the gradient and Hessian matrix of the intermediate activation in ANN.
Note that for squared error there is no third-order term. 
Since $\mathbf{g}^{(n)}$ is computed with respect to $\mathbf{x}^{(n)}$, it can be easily set to $\mathbf{0}$. Therefore, our task becomes to minimize $\mathbf{e}^{(n), \top}\mathbf{H}^{(n)}\mathbf{e}^{(n)}$. 

Then, we can rewrite the error term by:
\begin{align}
   \mathbf{e}^{(n)} & = \mathbf{x}^{(n)}-\bar{\mathbf{s}}^{(n)} \nonumber\\ 
   & = \mathrm{ReLU}(\mathbf{W}^{(n-1)}\mathbf{x}^{(n-1)}) - \mathrm{ClipFloor}(\mathbf{W}^{(n-1)}\bar{\mathbf{s}}^{(n-1)}) \nonumber\\
   & = \mathrm{ReLU}(\mathbf{W}^{(n-1)}\mathbf{x}^{(n-1)}) - \mathrm{ReLU}(\mathbf{W}^{(n-1)}\bar{\mathbf{s}}^{(n-1)}) \nonumber\\
   & + \mathrm{ReLU}(\mathbf{W}^{(n-1)}\bar{\mathbf{s}}^{(n-1)}) - \mathrm{ClipFloor}(\mathbf{W}^{(n-1)}\bar{\mathbf{s}}^{(n-1)}) \label{eq_expan_err}\\
   & = \mathbf{e}_r^{(n)} + \mathbf{e}_c^{(n)},
   \label{eq_split_err}
\end{align}
where the error term $\mathbf{e}_r^{(n)}$ containing the first two terms in \autoref{eq_expan_err} is the difference caused by different input activation, \ie $\mathbf{x}^{(n-1)}$ and $\bar{\mathbf{s}}^{(n-1)}$. The error term $\mathbf{e}_c^{(n)}$, which contains the last two terms in \autoref{eq_expan_err} is the difference caused by different activation function in current layer, \ie $\mathrm{ReLU}$ and $\mathrm{ClipFloor}$. Here, we will show that we can eliminate $\mathbf{e}_r^{(n)}$ with the following theorem.

\begin{theorem}
\label{theorem}
The conversion error in the final network output space can be bounded by a weighted sum of local conversion error, given by
\begin{equation}
\mathbf{e}^{(n), \top}\mathbf{H}^{(n)}\mathbf{e}^{(n)} \le \sum_{\ell=1}^n 2^{n-\ell+1} \mathbf{e}_c^{(\ell), \top}\mathbf{H}^{(\ell)}\mathbf{e}_c^{(\ell)}.
\end{equation}
\end{theorem}
We put our detailed proof of this theorem in \aref{append_proof}. A major ingredient of this proof is the recursive formulation of the Hessian matrix defined in \cite{botev2017practical}.

As stated by the above theorem, the distance at the final output space can be upper bounded by the activation distances ($\mathbf{e}_c$) in each local layer. Moreover, the shallow layer has a cumulative effect on the final layer output. 
In practice, the Hessian matrix in each layer $\mathbf{H}^{(\ell)}$ is intractable to compute and store. For instance, a 10 MB activation needs 100 TB of space to store the Hessian matrix. Therefore, following former practice~\cite{li2021brecq} we approximate the Hessian matrix as a constant diagonal matrix. Then, our final optimization objective is transformed to 
\begin{equation}
    \min \sum_{\ell=1}^n 2^{n-\ell+1} || \mathbf{e}_c^{(\ell)} ||_F^2.
\end{equation}

\begin{figure}[t]
 \centering
 \includegraphics[width=0.8\linewidth]{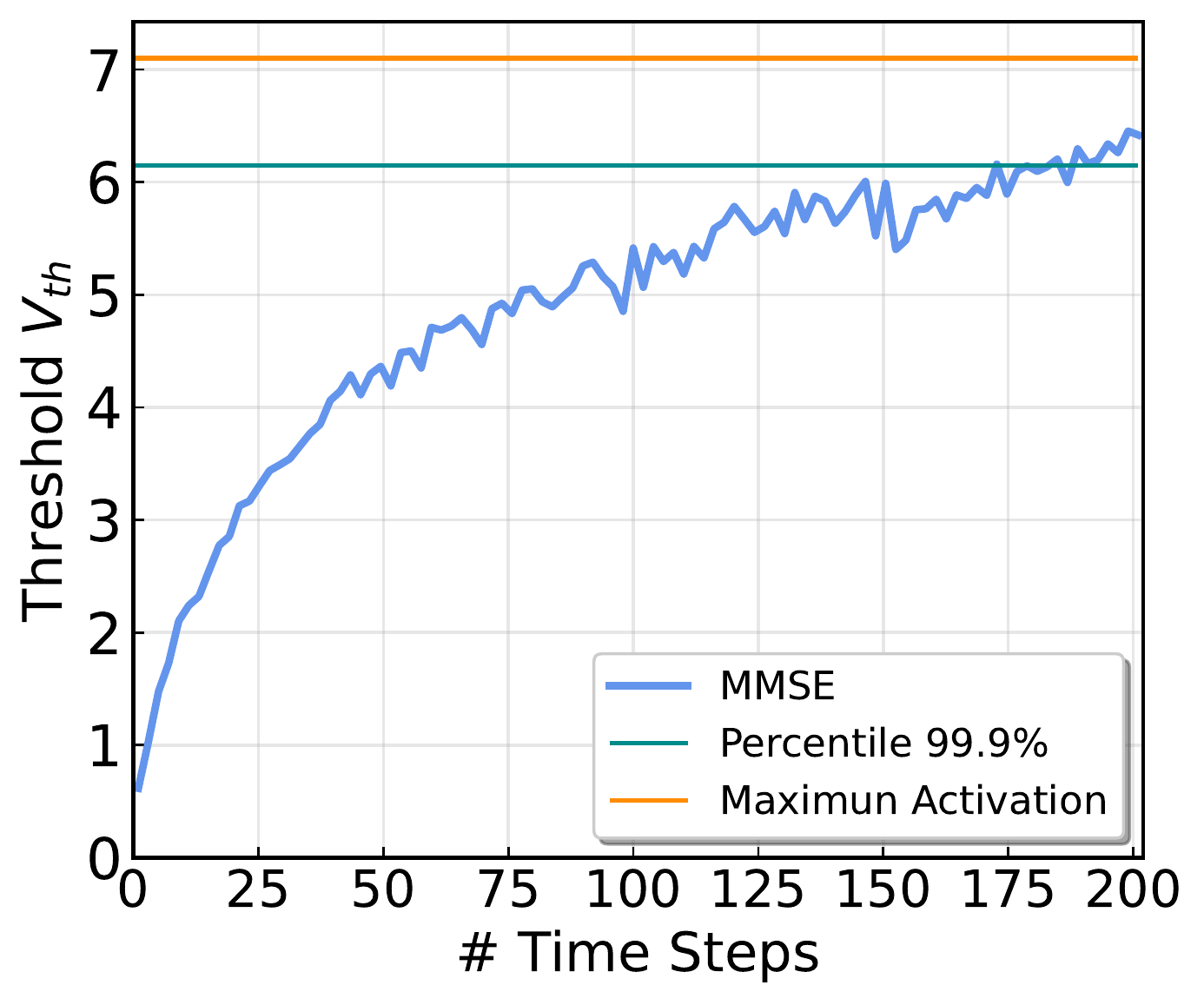}
 \caption{Comparison of the threshold determined by three different approaches. Our MMSE threshold is adaptive to different number of time steps.}
 \label{fig_mse}
\end{figure}

\subsection{ClipRound and Adaptive Threshold}
\label{sec_cr_at}
Our mission is to reduce the activation discrepancy in each layer. Here, we propose two simple adjustments that can efficiently and effectively reduce the local error. 

First, we suggest to use $\mathrm{ClipRound}$ function rather than $\mathrm{ClipFloor}$, where rounding a number $x$ is equal to flooring $x+\frac{1}{2}$. 
Original $\mathrm{ClipFloor}$ always produces positive error since $\mathrm{ReLU}(x)\le\mathrm{ClipFloor}(x)$. Using $\mathrm{ClipRound}$ the error includes both positive and negative values, which achieves a better balance. 
For input from an uniform distribution, $\mathrm{ClipRound}$ can achieves lowest flooring (or rounding) error~\cite{deng2021optimal}, given by
\begin{align}
   \bar{\mathbf{s}}^{(\ell+1)} & = \mathrm{ClipRound}(\mathbf{W}^{(\ell)}\bar{\mathbf{s}}^{(\ell)}, T, V_{th}^{(\ell)}) \nonumber\\ 
   & = \frac{V_{th}^{(\ell)}}{T}\mathrm{Clip}\left(\left\lfloor\frac{T}{V_{th}^{(\ell)}}\mathbf{W}^{(\ell)}\bar{\mathbf{s}}^{(\ell)}+\frac{1}{2}\right\rfloor, 0, T\right).
   \label{eq_clip_round}
\end{align}
In practice, we don't need to modify the IF neurons mechanism to fulfill the rounding operation. Rather, we could add $\nicefrac{V^{(\ell)}_{th}}{2T}$ to the bias parameter without additional overhead. 

Second, we propose to determine the firing threshold by leveraging the $\mathrm{ClipRound}$ function.
In an effort to better balance the rounding error and the clipping error, we use Minimization of Mean Squared Error (MMSE) to obtain the threshold $V_{th}^{(\ell)}$ given different pre-defined time step $T$, which is formulated by
\begin{equation}
    \min_{V_{th}} \left(\mathrm{ClipRound}(\bar{\mathbf{s}}^{(\ell+1)}, T, V_{th}^{(\ell)}) - \mathrm{ReLU}(\bar{\mathbf{s}}^{(\ell+1)})\right)^2
\end{equation}
Note that the above problem is not convex, therefore no closed-form solution is guaranteed. We hereby sample several batches of training images and use grid search to determine the final result of $V_{th}^{(\ell)}$. Specifically, we linearly sample $N$ grids between $[0, \max({\bar{\mathbf{s}}}^{(\ell+1)})]$, and find the grid with the lowest MSE. The detailed algorithm can be found in \autoref{alg_thresh}. In our implementation, we set $N=100$ and find this option achieves a good trade-off between search efficiency and precision.
\autoref{fig_mse} shows the dynamics of our proposed method. It is worthwhile to note that $V_{th}^{(\ell)}$ does not monotonically increase along with $T$, because the rounding error may be decreased by slightly incrementing the threshold. We can further apply the MMSE threshold channel-wisely to further decrease the MSE error, as inspired from~\cite{kim2019spiking}.

\begin{algorithm}[t]
    \caption{Searching the Firing Threshold }
    \label{alg_thresh}
    \KwIn{Pretrained ANN; Several training samples, grid size $N$.} 
    \For{all $i=1,2,\dots, n$-th layer in the ANN model}
            {
            Collect ANN output $\mathbf{x}^{(i)}$;\\
            Compute maximum activation $\max(\mathbf{x}^{(i)})$;\\
            \For{all $j=1,2,\dots,N$-th grid}
            { 
                Temporarily set threshold to $\frac{j}{N}\max(\mathbf{x}^{(i)})$;\\
                Compute MSE between ReLU and ClipFloor activations;\\ 
            }
            Find the threshold with minimum MSE;\\
            }
    \textbf{return} Thresholds in all IF neuron layers\;
\end{algorithm}

\subsection{Layer-wise Parameter Calibration}
\label{sec_calibration}
In \autoref{sec_err_analysis}, we showcase that in each layer, the total conversion error is composed of activation distance error and input distance error which can be decomposed to the former layers. Our analysis shows that the former layer has a cumulative effect on the final layer. 
In this section, we argue that we can view the conversion as an \textbf{optimization problem} to reduce the conversion error by \textbf{calibrating} the parameters, instead of simply copy-pasting the parameters from ANNs to SNNs. To facilitate fast calibration, we optimize the parameters in a layer-wise, greedy manner: We gradually calibrate the parameters by minimizing the activation distance error in one layer, starting with the first convolutional layer and continuing the calibration layer-by-layer. Thus, the cumulative effect from the former layer is reduced when we calibrate later layers. 
Specifically, we calibrate each layer's three parameters: bias $\mathbf{b}$, weights $\mathbf{W}$, and the initial membrane potential $\mathbf{v}(0)$.
Calibrating these parameters requires different computing and memory resources. Therefore, we introduce the Light Calibration Pipeline~(LCP) and the Advanced Calibration Pipeline~(ACP), which can be chosen by users according to their memory and computation budgets in practice.
The light calibration pipeline achieves fast calibration with less memory and computation consumption by only adjusting bias in SNNs. With a little effort, the light pipeline can outperform state-of-the-art methods by a large margin. The advanced calibration pipeline can achieve best results by calibrating the weights as well as the initial membrane potential in a fine-grained way. 

\vspace{2mm}
\noindent\textbf{Light Calibration Pipeline (LCP).} LCP only calibrate the bias parameters. To calculate the difference in bias, we first define a reduced mean function:
\begin{equation}
    \mu_i(\mathbf{x})= \frac{1}{wh}\sum_{i=1}^{w}\sum_{j=1}^{h} \mathbf{x}_{c,i,j}
\end{equation}
where $w, h$ are the width and height of the feature-map, and $\mu_i(\mathbf{x})$ computes the spatial mean of the feature-map in the $i$-th channel.
Note that the spatial mean of conversion error $\mathbf{e}_c^{(\ell)} = $ can be written by
\begin{equation}
   \mu_c(\mathbf{x}^{(\ell)}) = \mu_c(\bar{\mathbf{s}}^{(\ell)}) + \mu_c(\mathbf{e}_c^{(\ell)}). \label{eq_bc}
\end{equation}
To ensure the spatial mean of output in SNN is equal to the spatial mean of output in ANN, we can add the expected conversion error into the bias term as $\mathbf{b}^{(\ell)}_i:=\mathbf{b}^{(\ell)}_i+\mu_i(\mathbf{e}^{(\ell+1)})$. In practice, we only sample one batch of training images and compute the reduced mean to calibrate the bias and the LCP can be finished in seconds on GPUs (in minutes even on CPUs). The algorithm of LCP can be found in \autoref{alg_calibrate}.

\vspace{2mm}
\noindent\textbf{Advanced Calibration Pipeline (ACP).}
Calibrating the bias only corrects partial error. To further reduce conversion error, we propose an advanced calibration pipeline that consists of Potential Calibration (PC) and Weights Calibration (WC).

Now consider a non-zero initial membrane potential $\mathbf{v}^{(\ell)}(0)$, we can rewrite \autoref{eq_clip_round} to 
\begin{equation}
      \tilde{\bar{\mathbf{s}}}^{(\ell+1)} = \frac{V_{th}^{(\ell)}}{T}\mathrm{clip}\left(\left\lfloor\frac{T}{V_{th}^{(\ell)}}\mathbf{W}^{(\ell)}\bar{\mathbf{s}}^{(\ell)}+\frac{\mathbf{v}^{(\ell)}(0)}{V_{th}^{(\ell)}}+\frac{1}{2}\right\rfloor, 0, T\right),
\end{equation}
where $\tilde{\bar{\mathbf{s}}}^{(\ell+1)}$ is the calibrated expected output with non-zero initialization of membrane potential.
To obtain a fast calibration for the initial membrane potential, here we make an approximation to get a closed-form solution:
\begin{align}
    \tilde{\bar{\mathbf{s}}}^{(\ell+1)} & \approx \frac{V_{th}^{(\ell)}}{T}\mathrm{clip}\left(\left\lfloor\frac{T}{V_{th}^{(\ell)}}\mathbf{W}^{(\ell)}\bar{\mathbf{s}}^{(\ell)}+\frac{1}{2}\right\rfloor, 0, T\right)+\frac{\mathbf{v}^{(\ell)}(0)}{T} \nonumber \\
    & = \bar{\mathbf{s}}^{(\ell+1)} + \mathbf{v}^{(\ell)}(0)/T 
\end{align} 
Similar to bias calibration, $\mathbf{v}^{(\ell)}(0)/T$ can correct the output distribution of SNN. We can directly set $\mathbf{v}^{(\ell)}(0)$ to $T\times \mathbf{e}^{(\ell+1)}$ to calibrate the initial potential. 
Note that the potential calibration does not need to compute the spatial mean.

In ACP, we also introduce weights calibration to correct the conversion error in each layer. In weights calibration, the objective is given by:
\begin{align}
   \min_{\mathbf{W}^{(\ell)}}  ||\mathbf{e}_c^{(\ell+1)}||^2_F = \min_{\mathbf{W}^{(\ell)}} ||\mathbf{x}^{(\ell+1)}-\mathrm{ClipRound(\mathbf{s}^{(\ell)})}||^2_F \label{eq_wc}.
\end{align}
Here we optimize weights in SNN layer-by-layer to reduce the conversion error. For practical implementation, we will first store input samples in ANN $\mathbf{x}^{(\ell)}$ and input spikes samples in SNN of each time step $\mathbf{s}^{(\ell)}(t)$ and then compute the expected spike input by $\bar{\mathbf{s}}^{(\ell)}=\sum_{t=1}^T\mathbf{s}^{(\ell)}(t)$. To further compute the gradient of $\mathrm{ClipRound}$ function, we apply the Straight Through Estimator~\cite{bengio2013estimating} of the round operation which is done in some quantization works~\cite{rastegari2016xnor,li2019additive}, \ie
\begin{equation}
    \frac{\partial \lfloor x \rfloor}{\partial x} = 1.
\end{equation} 
To this end, we can use stochastic gradient descent to optimize weights.
During calibration for weights, the optimization process is very efficient compared to other direct training methods. This is because we first store the expected input from previous layers, and we do not have to perform $T$ times convolution like direct training methods. The major bottleneck of WC is storing the input of SNN. For example, if we set $T=1024$, then we will do 1024 times forwarding pass for one batch and accumulate them to get the final expected results. We summarize our overall workflow in \autoref{alg_calibrate}.

\begin{algorithm}[b]
    \caption{The Workflow of SNN Calibration.}
    \label{alg_calibrate}
    \KwIn{Pretrained ANN; Several training samples.} 
    \vspace{2mm}
    Fold BN Layers into Conv Layers (cf. \autoref{eq_bn_fold});\\
    Determine the optimal threshold in each layer using \autoref{alg_thresh};\\
    
    \For{all $i=1,2,\dots, n$-th layer in the ANN model}
            {
            Collect ANN input data $\mathbf{x}^{(i-1)}$, output data $\mathbf{x}^{(i)}$;\\
            Collect SNN input data $\bar{\mathbf{s}}^{(i-1)}$;\\
            Compute Error term $\mathbf{e}_c^{(i)}=\mathbf{x}^{(i)}-\bar{\mathbf{s}}^{(i)}$;\\
            \If{use LCP}{
             Calibrate bias term $\mathbf{b}^{(i)}:=\mathbf{b}^{(i)}+\mu(\mathbf{e}^{(i)})$;\\
            }
            \If{use ACP}{
            Calibrate potential $\mathbf{v}^{(i)}(0) := T\times\mathbf{e}^{(i)}$ ;\\
            Optimize weights to minimize $||\mathbf{e}^{(\ell)}||^2$  via stochastic gradient descent;\\
            }
            }
    \textbf{return} the calibrated SNN\;
\end{algorithm}

\begin{table*}[h]
 \caption{Ablating the design choices of our methods, including threshold determination, $\mathrm{ClipRound}$, parameter calibration pipelines. 
 We test the VGG16 and ResNet-20 on the ImageNet dataset.
 MaxAct~\cite{Diehl2015Fast} and Percentile~\cite{rueckauer2017conversion} are existing methods. 
 * denotes channel-wise threshold. Each result is averaged over 5 runs with different random seeds. }
    \begin{subtable}[h]{0.325\textwidth}
    \centering
    \footnotesize
    \caption{Accuracy comparison on SNN conversion (with or without $\mathrm{ClipRound}$) for ANNs \textbf{without} BN.} \label{tab_clipround}
    \centering
    \begin{adjustbox}{max width=1\textwidth}
    \begin{tabular}{lcccc}
    \toprule
    \multirow{2}{6em}{\textbf{Method}} & \multicolumn{2}{c}{\textbf{VGG-16 (72.43)}} & \multicolumn{2}{c}{\textbf{ResNet-34 (70.95)}} \\
    \cmidrule(l{2pt}r{2pt}){2-3}\cmidrule(l{2pt}r{2pt}){4-5}
    & T=16 & T=32 & T=16 & T=32  \\
    \midrule
    MaxAct$+\mathrm{ClipFloor}$ & 0.1 & 0.1 & 0.46 & 5.43  \\
    MaxAct$+\mathrm{ClipRound}$ & 44.80 & 63.22 & 7.46 & 44.60\\
    \midrule
    MMSE$+\mathrm{ClipFloor}$ & 9.53 & 67.11 & 5.54 & 51.18 \\
    MMSE$+\mathrm{ClipRound}$ & 68.86 & 71.08 & 28.70 & 60.50 \\
    \bottomrule
    \end{tabular}
    \end{adjustbox}
    \end{subtable}
    \hfill
    \begin{subtable}[h]{0.34\textwidth}
    \centering
    \caption{Accuracy comparison on different threshold determination methods for conversion from ANN \textbf{without} BN to SNN.} \label{tab_mmse_thresh}
    \begin{adjustbox}{max width=\textwidth}
    \begin{tabular}{lcccc}
    \toprule
    \multirow{2}{6em}{\textbf{Method}} & \multicolumn{2}{c}{\textbf{VGG-16 (72.43)}} & \multicolumn{2}{c}{\textbf{ResNet-34 (70.95)}} \\
    \cmidrule(l{2pt}r{2pt}){2-3}\cmidrule(l{2pt}r{2pt}){4-5}
    & T=16 & T=32 & T=16 & T=32 \\
    \midrule
    Maximum Act & 44.80 & 63.22 & 7.46 & 44.60 \\
    Percentile 99.99\% & 62.34 & 69.29 & 15.09 & 56.41 \\
    MMSE (Ours) & 68.86 & 71.08 & 28.70 & 60.50 \\
    MMSE* (Ours) & 68.92 & 71.23 & 43.69 & 60.35 \\
    \bottomrule
    \end{tabular}
    \end{adjustbox}
    \end{subtable}
    \hfill
    \begin{subtable}[h]{0.30\textwidth}
    \centering
    \footnotesize
    \caption{Accuracy comparison SNN conversion (use $\mathrm{ClipRound}$ and MMSE threshold) for ANN \textbf{with or without} BN.} \label{tab_bn}
    \centering
    \begin{adjustbox}{max width=1\textwidth}
    \begin{tabular}{lcccc}
    \toprule
    \multirow{2}{6em}{\textbf{Method}} & \multicolumn{2}{c}{\textbf{VGG-16}} & \multicolumn{2}{c}{\textbf{ResNet-34}} \\
    \cmidrule(l{2pt}r{2pt}){2-3}\cmidrule(l{2pt}r{2pt}){4-5}
    & T=16 & T=32 & T=16 & T=32 \\
    \midrule
    ANN without BN & 72.43 & 72.43 & 70.95 & 70.95  \\
    SNN & 68.86 & 71.08 & 28.70 & 60.50 \\
    \midrule
    ANN with BN & 75.66 & 75.66 & 75.36 & 75.36 \\
    SNN & 0.19 & 0.50 & 0.10 & 2.88 \\
    \bottomrule
    \end{tabular}
    \end{adjustbox}
    \end{subtable}
     
    \bigskip 
    
    \begin{subtable}[h]{0.47\textwidth}
    \centering
    \caption{Accuracy comparison by using LCP to different threshold methods on ANN \textbf{with BN} layers. } \label{tab_bias_corr}
    \begin{adjustbox}{max width=\textwidth}
    \begin{tabular}{lcccc}
    \toprule
    \multirow{2}{6em}{\textbf{Method}} & \multicolumn{2}{c}{\textbf{VGG-16 (75.66)}} & \multicolumn{2}{c}{\textbf{ResNet-34 (75.36)}} \\
    \cmidrule(l{2pt}r{2pt}){2-3}\cmidrule(l{2pt}r{2pt}){4-5}
    & T=16 & T=32 & T=16 & T=32 \\
    \midrule
    MaxAct + LCP & 0.94 & 1.18 & 0.99 & 1.55 \\
    Percentile 99.99\% + LCP & 1.34 & 3.01 & 0.11 & 4.57 \\
    MMSE + LCP (Ours) & 11.95 & 29.48 & 19.75 & 53.16 \\
    MMSE* + LCP (Ours) & \textbf{40.81} & \textbf{57.14} & \textbf{28.20} & \textbf{55.16} \\
    \bottomrule
    \end{tabular}
    \end{adjustbox}
    \end{subtable}
    \hfill
    \begin{subtable}[h]{0.47\textwidth}
        \centering
    \caption{Accuracy comparison by using ACP to different threshold methods on ANN \textbf{with BN} layers.} \label{tab_heavy_pipe}
    \begin{adjustbox}{max width=\textwidth}
    \begin{tabular}{lcccc}
    \toprule
    \multirow{2}{6em}{\textbf{Method}} & \multicolumn{2}{c}{\textbf{VGG-16 (75.66)}} & \multicolumn{2}{c}{\textbf{ResNet-34 (75.36)}} \\
    \cmidrule(l{2pt}r{2pt}){2-3}\cmidrule(l{2pt}r{2pt}){4-5}
    & T=16 & T=32 & T=16 & T=32 \\
    \midrule
    MaxAct + ACP & 0.10 & 1.70 & 0.41 & 39.60 \\
    Percentile 99.99\% + ACP & 0.13 & 52.24 & 14.24 & 58.13 \\
    MMSE + ACP (Ours) & 64.25 & 68.20 & 50.84 & \textbf{66.97} \\
    MMSE* + ACP (Ours) & \textbf{65.02} & \textbf{69.04} & \textbf{51.67} & 64.65 \\
    \bottomrule
    \end{tabular}
    \end{adjustbox}
    \end{subtable}
     \label{tab:temps}
\end{table*}

\subsection{Converting BN layers}
Batch normalization layers utilize linear transformation to each channel of activation after the convolution. It estimates the activation statistics with an exponential moving average. 
In SNN, the BN layers are hard to estimate the statistics in both spatial and temporal dimensions, therefore, we fold the BN parameters $(\mu, \sigma, \gamma, \beta)$ to the preceding layer's weight and bias $(\mathbf{W}, \mathbf{b})$~\cite{rueckauer2017conversion}:
\begin{equation}
    \mathbf{W}:= \mathbf{W}\frac{\gamma}{\sigma}, \ \ \ \mathbf{b}:= \beta + (\mathbf{b}-\mu)\frac{\gamma}{\sigma}, \label{eq_bn_fold}
\end{equation}
where $\mu, \sigma$ are the running mean and standard deviation, and $\gamma, \beta$ are the transformation parameters in the BN layer. After folding, we apply calibration to the BN-folded parameters.

\newcommand{\mypm}[1]{\color{gray}{\tiny{$\pm$#1}}}
\section{Experiments}

To demonstrate the effectiveness and the efficiency of our proposed algorithm, we conduct comprehensive and large-scale experiments on both image classification and object detection tasks with extremely low simulation length (\eg $T\le 256$). For classification, we test our algorithm on CIFAR~\cite{cifardataset} and the challenging ImageNet~\cite{deng2009imagenet} datasets. Beside regular network architectures (VGGs and ResNets), we test lightweight MobileNet~\cite{howard2017mobilenets}, neural architecture searched MNasNet~\cite{tan2019mnasnet}, and large-scale RegNetX~\cite{radosavovic2020regnet}. 
For detection tasks, we evaluate MS COCO~\cite{lin2014microsoft} with Faster R-CNN~\cite{ren2015faster} and RetinaNet~\cite{lin2017focal}. 

The organization of this section goes as follows: \autoref{sec_implement} briefly sorts out our implementation details. 
In \autoref{sec_ab_study}, we conduct ablation studies on the approximations and design choices made in~\autoref{sec_method}.
In \autoref{sec_sota_comp}, we compare our methods to state-of-the-art prior methods. We include the detection results in \autoref{sec_detect}. In \autoref{sec_vis}, we visualize the activation mismatch after our calibration algorithm. In \autoref{sec_complex}, we conduct a complexity study on space, time, and the size of the calibration dataset. 

\subsection{Implementation Details}
\label{sec_implement}
For all ANN with BN layers, we fold the BN layer before conversion. We do not convert input images to binary spikes because generating binary spikes requires time and degrades the accuracy~\cite{deng2021optimal}.
To correct the bias and membrane potential, we sample one batch of unlabeled data (128 training images). To estimate the MMSE threshold and calibrate weights, we use 1024 training images. 
We use Stochastic Gradient Descent with 0.9 momentum to optimize weights in weights calibration, followed by a cosine learning rate decay~\cite{loshchilov2016sgdr}. The learning rate is set to $10^{-5}$, and no L2 regularization is imposed. We optimize the weights in each layer with 5000 iterations. We will analyze the time and space complexity of our algorithm in the next section. The training details of ANN are included in the Appendix.

\subsection{Ablation Studies}
In this section, we verify the design choices of our proposed methods, including ClipRound spiking approximation, MMSE firing threshold, and layer-wise parameter calibration. In all ablation experiments, we test VGG-16~\cite{simonyan2014very} and ResNet-34~\cite{he2016deep} on the ImageNet ILSVRC-2012 dataset~\cite{deng2009imagenet}.
We run each experiment for 5 times with different random seeds and report the mean top-1 accuracy on the validation set. 

\label{sec_ab_study}

\noindent\textbf{Effect of ClipRound in Conversion.}
We first test the effectiveness of $\mathrm{ClipRound}$ (\ie adding an offset term in the bias as we show in \autoref{sec_cr_at}). 
In \autoref{tab_clipround}, we observe that $\mathrm{ClipRound}$ significantly improves the accuracy. For VGG-16, the $\mathrm{ClipRound}$ provides 63\% absolute accuracy improvement with 32 time steps. For ResNet-34, $\mathrm{ClipRound}$ brings more than 20\% absolute accuracy uplift with 16 time steps.

\noindent\textbf{Effect of MMSE Threshold in Conversion.}
We next study the effect of choosing different threshold $V_{th}$.
In \autoref{tab_mmse_thresh}, we show that maximum activation (threshold that uses the maximum activation value) has the lowest performance because of large flooring error.
Our MMSE threshold achieves better results than maximum using the maximum activation as threshold~\cite{Diehl2015Fast} and percentile threshold~\cite{rueckauer2017conversion} when $T =16$.
As an example, our method is 28.6\% higher in accuracy than percentile when converting ResNet-34 with 16 time steps.
We further consider the channel-wise MMSE threshold rather than a layer-wise one and find that channel-wise MMSE achieves slightly better accuracy.
The effect of channel-wise threshold is more significant in ANNs with BN layers which will be demonstrated in next paragraphs.

\newcommand{\cmark}{\color{ForestGreen}\ding{51}}%
\newcommand{\xmark}{\color{Red}\ding{55}}%

\begin{table*}[h]
    \footnotesize
        \caption{Conversion accuracy comparison between our algorithm with other existing SNN works on the ImageNet. \textit{Use BN} means using BN layers to optimize ANN. For our method, we run ``baselines" (channel-wise MMSE and $\mathrm{ClipRound}$ function) with or without parameter calibration. We report the mean accuracy and the standard deviation from 5 runs.}
        \centering
        \begin{adjustbox}{max width=\textwidth}
        \begin{tabular}{lcc lllllc}
        \toprule
        \textbf{Method} & \textbf{Use BN}  & \textbf{ANN Acc.} & \textbf{$T=16$} & \textbf{$T=32$} & \textbf{$T=64$} & \textbf{$T=128$} &\textbf{ $T=256$} & \textbf{ $T\ge 2048$} \\
        \midrule
        \multicolumn{9}{c}{\textbf{ResNet-34~\cite{he2016deep}    ImageNet}} \\
        \midrule
        Spike-Norm~\cite{Sengupta2018Going} & \xmark & 70.69 & - & - &- &- &- & 65.47 \\ 
        Hybrid Train~\cite{rathi2019enabling} & \xmark & 70.20 & - & - &- &- & 61.48 & 65.10 \\
        RMP~\cite{han2020rmp} & \xmark & 70.64 & - & - & - & - & 55.65 & 69.89 \\
        TSC~\cite{han2020deep} & \xmark & 70.64 & - & - & - & - & 55.65 & 69.93 \\
        Ours (Baseline) & \xmark & 70.95& 43.69\mypm{0.33}  & 60.33\mypm{0.29} & 66.24\mypm{0.11} & 68.77\mypm{0.09} & 69.82\mypm{0.07} & 70.98 \\
        \textbf{Ours (Baseline + LCP)} & \xmark & 70.95 & \textbf{54.03\mypm{0.39}}  & \textbf{64.43\mypm{0.21}} & \textbf{68.00\mypm{0.13}} & \textbf{69.49\mypm{0.11}} & \textbf{70.25\mypm{0.05}} & 70.97 \\
        Ours (Baseline) & \cmark & 75.66 & 0.22\mypm{0.06} & 10.83\mypm{0.67} & 52.45\mypm{0.28} & 69.03\mypm{0.07} & 73.38\mypm{0.05} & 75.08 \\
        \textbf{Ours (Baseline + LCP)} & \cmark & 75.66 & \textbf{28.20\mypm{1.25}}  & \textbf{55.16\mypm{0.68}} & \textbf{67.56\mypm{0.10}} & \textbf{72.48\mypm{0.04}} & \textbf{74.53\mypm{0.04}} &  75.44 \\
        \textbf{Ours (Baseline + ACP)} & \cmark & 75.66 & \textbf{51.67\mypm{0.15}}  & \textbf{64.65\mypm{0.17}} & \textbf{71.30\mypm{0.08}} & \textbf{73.94\mypm{0.02}} & \textbf{75.00\mypm{0.04}} & 75.45 \\
        \midrule
        \multicolumn{9}{c}{\textbf{VGG-16~\cite{simonyan2014very} ImageNet}} \\
        \midrule
        Spike-Norm~\cite{Sengupta2018Going} & \xmark & 70.52 & -  & - &- &- &- & 69.96 \\ 
        Hybrid Train~\cite{rathi2019enabling} & \xmark & 69.35 & - & - &- &- & 62.73 & 65.19 \\
        RMP~\cite{han2020rmp} & \xmark & 73.49 & - & - & - & - & 48.32 & 73.09 \\
        TSC~\cite{han2020deep} & \xmark  & 73.49 & - & - & - & - & 69.71 & 73.46 \\
        Ours (Baseline) & \xmark  & 72.40 & 68.76\mypm{0.14} & 70.20\mypm{0.04} & 71.06\mypm{0.05} & 71.61\mypm{0.04} & 71.28\mypm{0.06}  & 72.09 \\
        \textbf{Ours (Baseline + LCP)} & \xmark & 72.40 & 69.42\mypm{0.11} & \textbf{70.50\mypm{0.03}} & \textbf{71.29\mypm{0.05}} & \textbf{71.65\mypm{0.03}} & \textbf{71.78\mypm{0.07}} & 72.29 \\
        Ours (Baseline) & \cmark & 75.36 & 0.95\mypm{0.29} & 5.52\mypm{1.93} & 42.13\mypm{3.19} & 65.63\mypm{1.07} & 72.35\mypm{0.27} & 73.88 \\
        \textbf{Ours (Baseline + LCP)} & \cmark & 75.36 & \textbf{40.81\mypm{1.09}} & \textbf{57.14\mypm{1.83}} & \textbf{68.60\mypm{0.64}} & \textbf{72.54\mypm{0.22}} & \textbf{74.19\mypm{0.09}} & 75.15 \\
        \textbf{Ours (Baseline + ACP)} & \cmark & 75.36 & \textbf{65.02\mypm{0.37}} & \textbf{69.04\mypm{0.29}} & \textbf{72.52\mypm{0.13}}  & \textbf{74.11\mypm{0.05}}  & \textbf{74.75\mypm{0.03}} & 75.32 \\
        \midrule
        \multicolumn{9}{c}{\textbf{MobileNet~\cite{Howard2017MobileNetsEC} ImageNet}} \\ 
        \midrule
        Ours (Baseline) & \cmark & 73.40 & 0.10\mypm{0.01}  & 0.10\mypm{0.00} & 0.34\mypm{0.03}  & 1.99\mypm{0.12} & 22.46\mypm{1.26} & 68.21 \\
        \textbf{Ours (Baseline + LCP)} & \cmark & 73.40 & 0.10\mypm{0.00} & 0.20\mypm{0.02} & \textbf{15.47\mypm{0.95}} & \textbf{55.95\mypm{0.49}} & \textbf{66.35\mypm{0.23}} & \textbf{72.19}\\
        \textbf{Ours (Baseline + ACP)} & \cmark & 73.40 & 9.15\mypm{0.17} & \textbf{34.69\mypm{0.36}} & \textbf{54.51\mypm{0.19}} & \textbf{64.58\mypm{0.10}} & \textbf{68.69\mypm{0.09}} & \textbf{72.38} \\ 
         \midrule
         \multicolumn{9}{c}{\textbf{MNasNet$\times2$\cite{tan2019mnasnet} ImageNet}} \\ 
         \midrule
         Ours (Baseline) & \cmark & 76.56 & 0.10\mypm{0.01} & 0.13\mypm{0.01} & 3.34\mypm{0.49} & 37.38\mypm{2.24} & 64.88\mypm{0.44} & 73.78 \\
        \textbf{Ours (Baseline + LCP)} & \cmark & 76.56 & 1.64\mypm{0.28} & \textbf{19.25\mypm{1.89}} & \textbf{52.53\mypm{0.51}} & \textbf{66.69\mypm{0.11}}  & \textbf{72.09\mypm{0.04}} & \textbf{74.84}\\
        \textbf{Ours (Baseline + ACP)} & \cmark& 76.56 & 5.34\mypm{0.75} & \textbf{44.56\mypm{1.68}} & \textbf{65.49\mypm{0.08}} & \textbf{70.30\mypm{0.11}} & \textbf{73.10\mypm{0.08}} & \textbf{75.10} \\ 
         \midrule
        \multicolumn{9}{c}{\textbf{RegNetX-4GF~\cite{radosavovic2020regnet} ImageNet}} \\ 
        \midrule
        Ours (Baseline) & \cmark & \textbf{80.07} & 0.14\mypm{0.01} & 0.22\mypm{0.02} & 3.25\mypm{0.15} & 47.95\mypm{0.30} & 71.14\mypm{0.16} &  \\
        \textbf{Ours (Baseline + LCP)} & \cmark & \textbf{80.07} & 0.68\mypm{0.05} & \textbf{28.04\mypm{0.66}} & \textbf{64.89\mypm{0.75}} & \textbf{74.10\mypm{0.10}}  & \textbf{77.20\mypm{0.05}} & \textbf{79.15}\\
        \textbf{Ours (Baseline + ACP)} & \cmark & \textbf{80.07} & \textbf{25.01\mypm{0.36}} & \textbf{57.10\mypm{0.11}} & \textbf{70.96\mypm{0.08}} & \textbf{75.23\mypm{0.06}} & \textbf{77.55\mypm{0.04}} & \textbf{79.21} \\
        \bottomrule
        \end{tabular}
        \end{adjustbox} \label{tab_sota_compare}
\end{table*}

\noindent\textbf{Prior Conversion Methods Performs Badly on ANNs with BN. }
We have shown that $\mathrm{ClipRound}$ and MMSE threshold can handle the conversion of ANN without BN layers. Here, we examine these two techniques on ANNs with BN. As shown in \autoref{tab_bn}, using BN is able to substantially improve the accuracy of ANNs, which are more ideal to be used for conversion. 
For example, VGG-16 with BN trained on ImageNet has 3.2\% higher accuracy than VGG-16 without BN. 
However, they suffer a large accuracy deficit when converting models with BN layers. The Spiking VGG-16 with BN easily crashed (0.5\% accuracy) even equipped with the $\mathrm{ClipRound}$ and the MMSE threshold techniques, and so does ResNet-34 (2.9\% accuracy). In both cases, the converted SNNs from ANNs with BN-layers exhibit extremely low convertibility.

\noindent\textbf{LCP: Combining Bias Calibration. }
Next, we verify the effect of the proposed parameter calibration by applying the light calibration pipeline (LCP) to different threshold methods. Results are summarized in \autoref{tab_bias_corr}, where we can find BC can \textit{consistently improve the accuracy of converted SNN by simply tuning the bias parameters.} For example, LCP boosts spiking VGG-16 to 36.6\% (from 0.2\%) accuracy using channel-wise MMSE threshold when $T=16$. With higher time steps, say 32, the LCP even achieves 56.6\% accuracy, a substantial improve compared with the methods without any calibration in \autoref{tab_bn}. We should emphasize that BC is cheap and only requires tiny memory space (\eg 0.36 MB for a ResNet-34) to store the bias term for different $T$.

\noindent\textbf{ACP: Potential and Weights Calibration. }
Finally, we verify the most advanced calibration pipeline, containing potential and weights calibration that will alter the ANN's parameters to adapt better in spiking configuration. We validate the effect of ACP on our MMSE threshold mode in \autoref{tab_heavy_pipe}. This time, the spiking VGG-16 can score 65\% accuracy even with 16 time steps. 
The spiking ResNet-34 obtains 50\% accuracy.
It is worthwhile to note that our layer-wise calibration algorithms should be combined with MMSE threshold. 
For example, the ResNet-34 still crashes with maximum activation threshold and ACP. 

\subsection{Comparison to Previous Work}
\label{sec_sota_comp}

\subsubsection{ImageNet}

In this section, we compare our proposed algorithm with other existing works evaluated on the image classification task. We first test ImageNet models. Here we choose the widely adopted ResNet-34~\cite{he2016deep} and VGG-16~\cite{simonyan2014very} in the existing literature. Note that we test ANNs both with and without BN layers. We additionally verify our algorithm on a compact architecture, {MobileNet}~\cite{Howard2017MobileNetsEC} and a neural architecture searched model, MNasNet~\cite{tan2019mnasnet}, and a large-scale network with over 80\% ImageNet accuracy, RegNetX-4GF~\cite{radosavovic2020regnet}. We compare the Spike-Norm, Hybrid Train, RMP, and TSC~\cite{Sengupta2018Going, rathi2019enabling, han2020rmp, han2020deep} as existing baselines. \emph{For our methods, we construct a baseline that uses channel-wise MMSE threshold and ClipRound and test this baseline method with light/advanced calibration pipelines.}

\begin{table*}[h]
\footnotesize
\caption{Conversion accuracy comparison between our algorithm with other existing SNN works on the CIFAR. \textit{Use BN} means using BN layers to optimize ANN. For our method, we run ``baselines" (channel-wise MMSE and $\mathrm{ClipRound}$ function) with or without parameter calibration. We report the mean accuracy and the standard deviation from 5 runs.}
\centering
\begin{adjustbox}{max width=\textwidth}
\begin{tabular}{lc | clllll | clllll}
\toprule
\multirow{2}{7em}{\textbf{Method}} & \textbf{Use} & \multicolumn{6}{c}{\textbf{CIFAR-10}} & \multicolumn{6}{c}{\textbf{CIFAR-100}} \\
\cmidrule(l{2pt}r{2pt}){3-8} \cmidrule(l{2pt}r{2pt}){9-14}
& \textbf{BN} & ANN  & \textbf{$T=4$} & \textbf{$T=8$} & \textbf{$T=16$} & \textbf{$T=32$} &\textbf{ $T\ge512$} 
& ANN & \textbf{$T=4$} & \textbf{$T=8$} & \textbf{$T=16$} & \textbf{$T=32$} &\textbf{ $T\ge512$}\\
\midrule
\multicolumn{14}{c}{\textbf{ResNet-20~\cite{he2016deep}}} \\
\midrule
Spike-Norm~\cite{Sengupta2018Going} & \xmark & 89.10 & - & - &- &- & 87.46 & 69.72 & - & - & - & - & 64.09 \\ 
Hybrid Train~\cite{rathi2019enabling} & \xmark & 93.15 & - & - &- &- & 92.94 & N/A & - & - &- &- & - \\
RMP~\cite{han2020rmp} & \xmark & 91.47 & - & - & - & - & 91.36 & 68.72 & - & - &- & 27.64 & 67.82 \\
TSC~\cite{han2020deep} & \xmark & 91.47 & - & - & - & 69.38 & 91.42 & 68.72 & - & - & - & 58.42 & 68.18 \\
Ours (Baseline) & \cmark & 96.72 & 27.21\mypm{3.62} &  69.48\mypm{2.88} & 89.49\mypm{0.27} & 94.71\mypm{0.20} & 90.72 & 81.51 & 17.01\mypm{0.90} & 32.59\mypm{2.26} & 67.50\mypm{0.54} & 77.42\mypm{0.39} & 81.28\\
\textbf{Ours (Baseline + LCP)} & \cmark & 96.72 & \textbf{75.07\mypm{0.88}} & \textbf{89.58\mypm{0.17}} & \textbf{94.77\mypm{0.12}} & \textbf{96.23\mypm{0.06}} & 96.75 & 81.51 & \textbf{47.61\mypm{1.61}} & \textbf{66.74\mypm{0.66}} & \textbf{76.38\mypm{0.34}} & \textbf{79.96\mypm{0.15}} & 81.30\\
\textbf{Ours (Baseline + ACP)} & \cmark & 96.72 & \textbf{84.70\mypm{0.71}} & \textbf{92.98\mypm{0.47}} & \textbf{95.51\mypm{0.06}} & \textbf{96.45\mypm{0.04}} & 96.74 & 81.51 & \textbf{54.96\mypm{1.11}} & \textbf{71.86\mypm{0.37}} & \textbf{78.13\mypm{0.25}} & \textbf{80.56\mypm{0.09}} & 81.39 \\
\midrule
\multicolumn{14}{c}{\textbf{VGG-16~\cite{simonyan2014very}}} \\
\midrule
Robust-Norm~\cite{rueckauer2017conversion} & \xmark & 92.82 & - & - & 10.11 & 43.03 & 92.75 & N/A & - & - & - & - & - \\ 
Spike-Norm~\cite{Sengupta2018Going} & \xmark & 91.70 & -  & - &- &- & 91.55 & 71.22 & - & - & - & - & 70.77 \\ 
Hybrid Train~\cite{rathi2019enabling} & \xmark & 92.81 & - & - &- &- & 92.48 & N/A & - & - & - & - & -  \\
RMP~\cite{han2020rmp} & \xmark & 93.63 & - & - & - & 60.30 & 93.63 & 71.22 & - & - & - & 63.76 & 70.93 \\
TSC~\cite{han2020deep} & \xmark  & 93.63 & - & - & - & - & 93.63 & 71.22 & - & -& - & 69.86 & 70.97 \\
RNL~\cite{ding2021optimal} & \xmark & 92.82 & - & - & 57.90 & 85.40 & 92.95 & N/A & - & - & - & - & - \\
Ours (Baseline) & \cmark & 95.60 & 30.56\mypm{6.89} & 71.91\mypm{4.08} & 83.45\mypm{0.92} & 90.91\mypm{0.72} & 95.53 & 77.93 & 5.97\mypm{0.93} & 30.12\mypm{2.02} & 44.00\mypm{1.50} & 61.97\mypm{1.02} & 77.45 \\
\textbf{Ours (Baseline + LCP)} & \cmark & 95.60 & \textbf{71.53\mypm{3.19}} & \textbf{90.36\mypm{0.50}} & \textbf{91.93\mypm{0.18}} & \textbf{94.38\mypm{0.15}} & 95.58 & 77.93 & \textbf{30.12\mypm{2.37}} & \textbf{57.51\mypm{0.25}} & \textbf{67.81\mypm{0.52}} & \textbf{73.59\mypm{0.28}} & 77.63 \\
\textbf{Ours (Baseline + ACP)} & \cmark & 95.60 & \textbf{86.57\mypm{1.54}} & \textbf{91.41\mypm{0.43}} & \textbf{93.64\mypm{0.09}} & \textbf{94.81\mypm{0.12}} & 95.60 & 77.93 & \textbf{55.60\mypm{1.55}} & \textbf{64.13\mypm{1.16}} & \textbf{72.23\mypm{0.50}} & \textbf{75.53\mypm{0.11}} & 77.79\\
\bottomrule
\end{tabular}
\end{adjustbox} \label{tab_cifar}
\end{table*}

Results can be found in \autoref{tab_sota_compare}. For both ResNet-34 and VGG-16 without BN, our baseline can generally achieve good performance. For example, converting VGG-16 with the baseline method only has a 1.3\% accuracy gap with ANN in 32 time steps. Our LCP, in this case, can slightly increase the conversion performance.
On models with BN layers, our method can substantially narrow the conversion loss. In particular, both VGG-16 and ResNet-34 crash at 16 time steps, however, our LCP can save 30~40\% accuracy. When applying the ACP, the VGG-16 even restores 65\% accuracy. As the time extends, the SNN converted from ANN with BN layers demonstrates higher potential: the 64 time steps spiking ResNet-34 converted with BN layers has 3.3\% higher accuracy than that converted without BN layers.

We further demonstrate the superiority of our algorithm by converting more sophisticated and compact neural architectures which could be sensitive to conversion. The MobileNet~\cite{howard2017mobilenets} is a representative efficient model with a small number of parameters and low amount of computation and MNasNet is a model obtained from automatic Neural Architecture Search~(NAS)~\cite{tan2019mnasnet}. As an example, our baseline methods fail to convert MobileNet at 128 time steps (only 2.2\% accuracy). In that case, our calibration algorithm is still able to recover the high performance. Using the ACP, the spiking MobileNet can have 64.6\% accuracy at 128 time steps, bringing a 62.4\% absolute accuracy improvement. Finally, we test our algorithm on a large-scale ANN, RegNetX-4GF~\cite{radosavovic2020regnet} which achieves 80.0\% top-1 accuracy with 22.1 million parameters. Our LCP reaches 74.1\% accuracy when $T=128$ and our ACP reaches 75.2\% accuracy when $T=128$.

\subsubsection{CIFAR10/100}
We also conduct experiments on the CIFAR dataset, using ResNet-20 and VGG-16 both with BN layers. From \autoref{tab_cifar}, we can observe that ANN trained with BN layers has significantly higher accuracy, which could also benefit SNN if we could appropriately convert the ANN to the SNN. We compare the existing literature the same with our ImageNet experiments. Notably, the baseline method fail to achieve good accuracy in low time steps $T=4$. For instance, the baseline spiking ResNet-20 only has 27.2\% accuracy on the CIFAR-10 dataset. Employing the ACP can drastically improve the performance of the SNN, increasing from 27.2\% to 84.7\%. The ResNet-20 reaches 95.5\% accuracy at 16 time steps, even higher than some direct training methods~\cite{zheng2020going}. For the CIFAR100 dataset, the conversion becomes more challenging. Our LCP and ACP still consistently output better conversion models. The VGG-16 increases from 6.0\% to 55.6\% accuracy at 4 time steps.

\begin{table}[t]
    \footnotesize
        \centering
        \caption{Comparison of mAP on COCO detection tasks among different calibration methods. We test RetinaNet~\cite{lin2017focal} and Faster R-CNN~\cite{ren2015faster} with ResNet-50 as the backbone. } \label{tab_detect}
        \begin{adjustbox}{max width=0.48\textwidth}
        \begin{tabular}{lrrrr}
        \toprule
        \textbf{Method} & $T=32$ & $T=64$ & $T=128$ & $T=256$ \\
        \midrule
        \multicolumn{5}{c}{\textbf{RetinaNet~\cite{lin2017focal}, Backbone: ResNet-50 (mAP: 35.6)}} \\
        \midrule
        Baseline & 6.2 & 20.1 & 28.7 & 32.0 \\
        Baseline + LCP & 19.2 & 27.4 & 31.2 & 33.0 \\
        Baseline + ACP & \textbf{28.4} & \textbf{32.3} & \textbf{33.7} & \textbf{34.3} \\
        \midrule
        \multicolumn{5}{c}{\textbf{Faster R-CNN~\cite{ren2015faster}, Backbone: ResNet50 (mAP: 37.0)}}\\
        \midrule
        Baseline & 16.6 & 27.6 & 32.3 & 34.4 \\
        Baseline + LCP & 25.5 & 31.0 & 33.7 & 35.0 \\
        Baseline + ACP & \textbf{31.0} & \textbf{34.2} & \textbf{35.4} & \textbf{35.9} \\
        \bottomrule
        \end{tabular}
        \end{adjustbox}
\end{table}

\subsection{Evaluation on Object Detection Tasks}
\label{sec_detect}

In this section, we extend our empirical experiments to MS COCO~\cite{lin2014microsoft} object detection tasks. 
We examine two classic detection algorithms: one-stage RetinaNet~\cite{lin2017focal} and two-stage Faster R-CNN~\cite{ren2015faster}, both of them use the ResNet-50 as backbone and Feature Pyramid Network~\cite{lin2017feature} as neck.
Pre-trained RetinaNet obtains 35.6 mean average precision (mAP) and pre-trained Faster R-CNN gets 37.0 mAP on the COCO dataset.
Note that we only convert the backbone to SNN, the accumulation of the pre-synaptic potential will be passed to the neck for further processing. 
Similar to the previous image classification experiments, we test baseline methods and optionally add calibration pipelines. The results are summarized in \autoref{tab_detect}.
Notably, the baseline method on RetinaNet degrades much performance at low time steps $(T=32)$, preserving only 6.2 mAP, yet our LCP can improve mAP to 19.2. 
The ACP even achieves 25.8 mAP within 32 time steps, which approaches the baseline result in 128 time steps. 
The Faster R-CNN seems to be more resilient than RetinaNet. The baseline method has 16.6 mAP with 32 time steps. Our ACP nearly doubles this performance. We can find that ACP can achieve higher performance with baseline methods while using only a half of the time steps.

\begin{figure*}
    \centering
    \includegraphics[width=\textwidth]{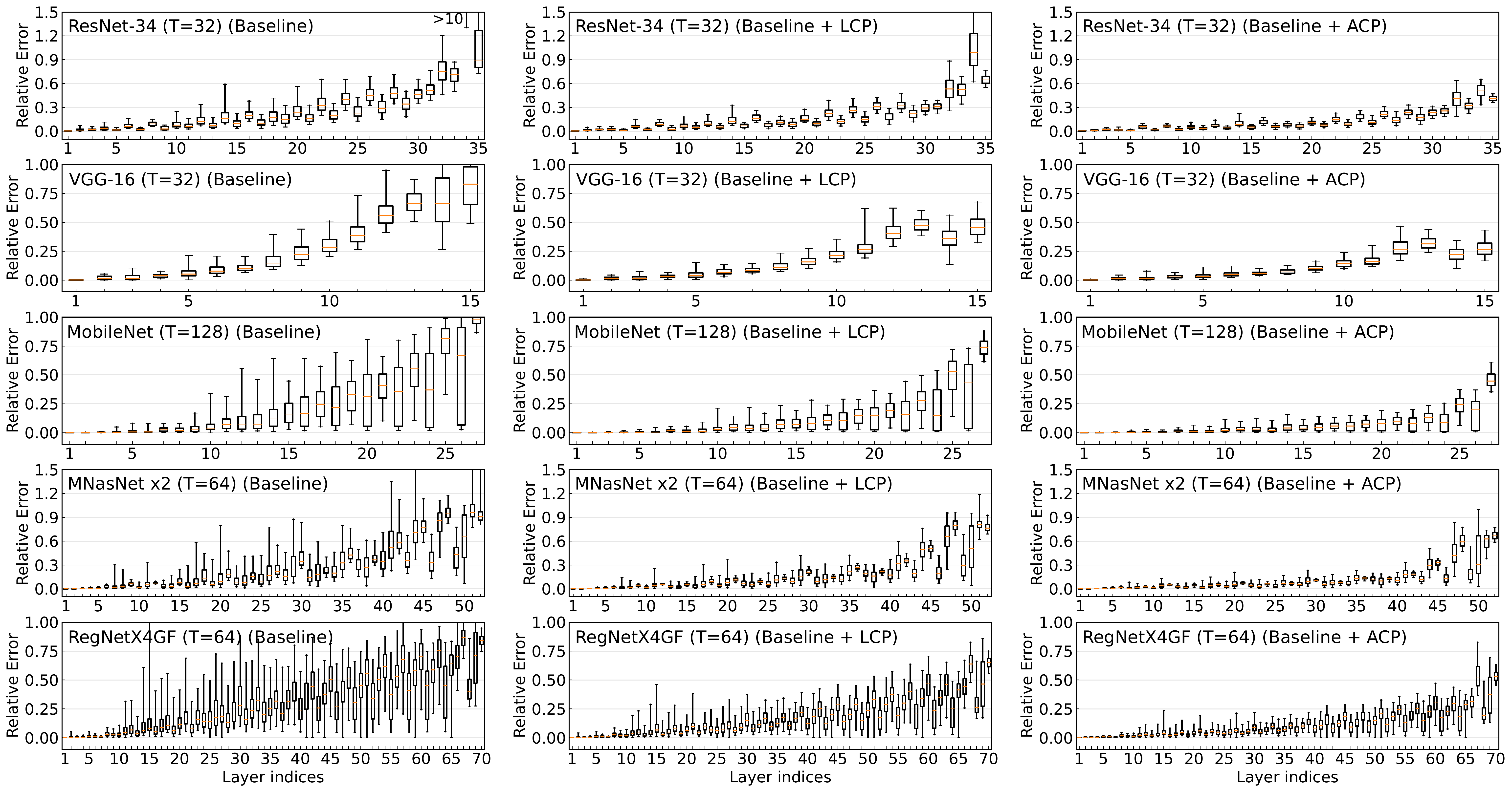}
    \caption{Measuring the relative error on the ImageNet models, all of which are trained with BN layers. We compare baseline and baseline with LCP/ACP, demonstrating the ability to reduce conversion error of our calibration algorithm. }
    \label{fig_relative_err_after_calibration}
\end{figure*}

\subsection{Visualization of Activation Mismatch}
\label{sec_vis}
As previously envisioned in \autoref{fig_relative_err}, we give a quantitative evaluation of activation mismatch (by using the \emph{relative error}), especially for the ANN with BN layers.  
In this section, we also plot the relative error using the baseline method and baseline with LCP or ACP. 
We select the 5 models on the ImageNet dataset and convert them to SNN with certain time steps.
We are concerned with whether our calibration pipelines are able to reduce the relative error of activation. The results of \autoref{fig_relative_err_after_calibration} are summarized as below. First, nearly all models have a near-lossless relative error (close to 0) in the first layer. However, the relative error steadily increases as the activation propagates throughout layers.
Eventually, at the last several layers, the relative error reaches $>1$, which means the error norm is even higher than the original activation norm. This observation confirms our analysis in \autoref{sec_err_analysis} that the conversion error will accumulate through the layers. Second, using our calibration algorithm, the activation mismatch is significantly ameliorated. Take ResNet-34 as an example, the baseline method shows a very high mismatch: the penultimate layer even has $>10$ relative error.
After ACP, the majority of the layers have lower than 0.3 relative error, and the penultimate layer only has 0.6 maximum relative error, \ie reduced to $\nicefrac{1}{17}$ of the original level.  
Last, we find that advanced architectures such as MobileNet and RegNet exhibit a higher variance of the relative error across channels. On RegNetX-4GF, we notice a significantly oscillated mismatch across channels. Again, our proposed parameter calibration algorithms effectively mitigate these problems. The first $\nicefrac{2}{3}$ layers' relative error is reduced under 0.25 using ACP.

\subsection{Complexity Study}

\begin{table}[t]
    \footnotesize
        \centering
        \caption{Conversion time (minutes) for different calibration algorithm. We set $T=64$ and test VGG-16 on CIFAR100 and MobileNet on ImageNet. ``m''= minutes, ``s''= seconds.}
        \begin{adjustbox}{max width=0.48\textwidth}
        \begin{tabular}{lccc}
        \toprule
        \textbf{Model} & \textbf{Bias}  & \textbf{Potential} & \textbf{Weights} \\
        \midrule
        VGG-16 (GPU) & 14.71s & 15.00s & 4m42s \\
        VGG-16 (CPU) & 3m56s & 3m54s & N/A \\
        MobileNet (GPU) & 2m17s & 2m11s & 27m36s \\ 
        MobileNet (CPU) & 33m13s & 33m57s & N/A \\ 
        \bottomrule
        \end{tabular}
        \end{adjustbox}
 \label{tab_convert_time}
\end{table}

\label{sec_complex}
\textbf{Time Complexity.}
During run-time, our converted SNN will not produce additional inference time compared to other conversion algorithms, and notably, doesn't require any modifications in ANN training. The conversion process itself may require time and computing resources. In \autoref{tab_convert_time}, we report the time needed for our calibration algorithms. 
All experiments were conducted on a single NVIDIA GTX 1080TI averaged from 5 runs.
We also test bias or potential calibration on a CPU using Intel XEON(R) E5-2620 v4. 
In \autoref{tab_convert_time}, we can see that the bias and potential calibration only takes a limited time, \eg 15 seconds on CIFAR100 and 4 minutes on ImageNet. This lightweight algorithm is affordable even on a single CPU. CPU can complete the bias calibration of VGG-16 on CIFAR100 within merely 4 minutes. This efficiency paves the way for deploying SNN on the edge IoT devices. 
Weights calibration is more expensive than the other two methods since for weights there are no closed-form solutions and thus weights have to rely on SGD optimization.
For example, the ACP for MobileNet on ImageNet may take 30 minutes on a GPU. However, it is still $\sim$200 times faster than Hybrid Train~\cite{rathi2019enabling}, which requires 20 epochs of end-to-end training (hundreds of GPU hours).

\begin{table}[t]
    \footnotesize
        \centering
        \caption{Comparison of the accuracy using different number of data samples for calibration.} \label{tab_num_sample}
        \begin{adjustbox}{max width=0.48\textwidth}
        \begin{tabular}{lllll}
        \toprule
        \textbf{\#Samples} & 32 & 64 & 128 & 256 \\
        \midrule
        ResNet-34 (LCP) & 57.23\mypm{0.83} & 62.46\mypm{0.29} & 65.09\mypm{0.30} & 66.23\mypm{0.18} \\
        \midrule
        \textbf{\#Samples} & 128 & 256 & 512 & 1024 \\
        \midrule
        ResNet-34 (ACP) & 69.73\mypm{0.12} & 69.80\mypm{0.13} & 70.05\mypm{0.15} & 70.57\mypm{0.14} \\ 
        \bottomrule
        \end{tabular}
        \end{adjustbox}
\end{table}

\vspace{2mm}
\noindent\textbf{Space Complexity. }
Here, we report the memory requirements for each calibration algorithm. Since our method will calibrate a new set of parameters for different $T$, therefore it is necessary to study the model size if we want to deploy SNNs under different $T$ settings. Specifically, calibrating the bias of ResNet-34 on ImageNet only requires 0.3653MB of memory. However, calibrating the weights and potential requires 83.25MB and 18.76MB, respectively. 
Thus, our proposed LCP is both computational and memory cheap and is optimal for flexible SNN conversion. In contrast, the ACP (including weights and potential calibration) requires more memory space. One may optionally only apply potential calibration to lower the memory footprint of ResNet-34. Interestingly, some tiny structures like MobileNet share less weight memory (12.21MB) but higher activation memory (19.81MB).

\vspace{2mm}
\noindent\textbf{Data Sample Complexity. }
We study the robustness of our algorithm by increasing the size of the dataset used for calibration. Here we test LCP with $32\sim256$ data samples and ACP with $128\sim1024$ data samples on ResNet-34 (ImageNet dataset). The results are put in \autoref{tab_num_sample}. By increasing the number of samples for calibration, the accuracy will also increase. 
We can find that more data samples are very effective on LCP. 32 training images only score 57.2\% accuracy while 256 training images obtain 67.5\% accuracy. Therefore, we recommend using at least 256 images for LCP. As for ACP, the robustness against the size of the calibration dataset is higher than LCP. Increasing the number of samples from 128 to 1024 only brings 0.8\% mean accuracy improvement. We also find that more samples lead to a stable calibration result, \ie less variance with more data samples. We need to emphasize that in practice collecting 1024 images is very trivial for a dataset like ImageNet, with approximately only 1 image per class. 

\subsection{Calibration without Training Data}

\begin{table}[t]
\footnotesize
\centering
\caption{Comparison of the accuracy using different source of calibration data.} 
\label{tab_fake_data}
\begin{adjustbox}{max width=0.48\textwidth}
\begin{tabular}{lllll}
\toprule
\textbf{Methods} & $T=16$ & $T=32$ & $T=64$ & $T=128$ \\
\midrule
 & \multicolumn{4}{c}{Original ImageNet Data} \\
\cmidrule{2-5}
\multirow{2}{8em}{ResNet-34 (ACP)} & \textbf{51.67\mypm{0.15}}  & \textbf{64.65}\mypm{0.17} & \textbf{71.30\mypm{0.08}} & \textbf{73.94\mypm{0.02}} \\
\cmidrule{2-5}
& \multicolumn{4}{c}{Distilled Fake Data~\cite{li2021mixmix}} \\
\cmidrule{2-5}
 & 41.25\mypm{0.32} & 62.32\textbf{\mypm{0.04}} & 70.11\mypm{0.09} & 73.20\mypm{0.04} \\ 
\bottomrule
\end{tabular}
\end{adjustbox}
\end{table}

In the case where the source training data is private or unavailable due to copyright issues, we can generate fake data by inverting the knowledge learned in the pre-trained ANN~\cite{kim2021privatesnn}. Doing so can remove the reliance on collecting calibration data from the training dataset. 

Here, we test an image synthesis technique~\cite{li2021mixmix} to create 1024 fake images and use them as our calibration dataset. 
We test ResNet-34 using the ACP algorithm in \autoref{tab_fake_data}. It can be seen that fake data have a larger gap with real data when the time step is low. When $T=16$, the real ImageNet data has a 10\% higher accuracy improvement. However, the gap could be significantly narrowed if we convert SNN in 128 time steps, \eg only 0.74\% accuracy difference.

\subsection{Efficiency and Sparsity}
\begin{figure}[t]
    \centering
    \includegraphics[width=0.4\textwidth]{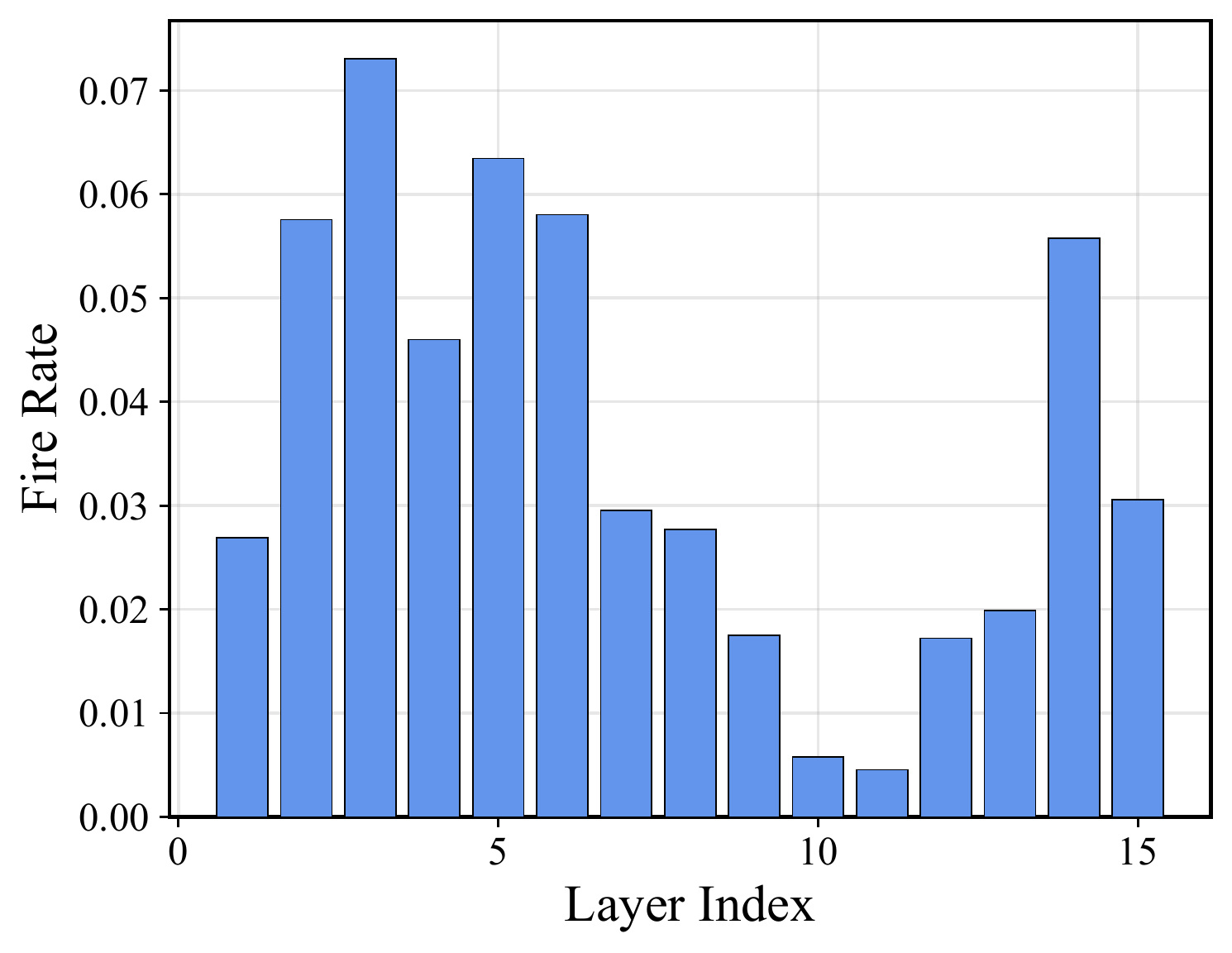}
    \caption{Firing rate visualization of VGG-16.}
    \label{fig_fire_rate}
\end{figure}

In this section, we visualize the sparsity of our calibrated SNN. We choose the spike VGG-16 on the ImageNet dataset, with $T=64$. We leverage the LCP (only bias calibration) and record the mean firing ratio across the whole validation dataset, which also corresponds to the sparsity of the activation. The firing ratio is demonstrated in \autoref{fig_fire_rate}, where we can find the maximum firing ratio is under 0.08, and the minimum firing ratio can be 0.025. To quantitatively compute the energy saving, we use the energy-estimation equation in~\cite{rathi2021diet}. For addition operation, we measure it by 0.9$J$ per operation; for multiplication, we measure it by 4.6$J$ per operation. On the event-driven neuromorphic hardware, a non-firing neuron will not cost any energy. Based on this rule, our calibrated spiking VGG-16 only costs 69.36\% energy of ANN's consumption.

\section{Conclusion}

In this work, we systematically analyze the error produced in the ANN-SNN conversion. We first demonstrate that the local error can be dissected into clipping error and flooring error. Then we derive the error propagation using the Hessian matrix information. Meanwhile, we showcase that ANNs armed with BN layers tend to act up due to severe activation mismatch. Both these theoretical and empirical findings motivate us to develop a greedy layer-wise parameter calibration algorithm. Different from the conventional copy-paste conversion algorithms, parameter calibration views the conversion as an optimization problem and adjusts the parameters accordingly to match to activation distribution in the source ANN. We further propose two pipelines that can adapt to different user requirements for computation and memory. 

We examine our calibration algorithms comprehending both image classification and object detection tasks, especially for large-scale datasets and modern neural architectures. 
Our method establishes a new state of the art for SNN conversion. It can successfully convert challenging architectures like MobileNet and RegNetX-4GF with low latency (less than 256 time steps) for the first time. Even when converting the ANN with Batch Normalization layers, our method can preserve high classification accuracy.


%

\ifCLASSOPTIONcompsoc
  \section*{Acknowledgments}
\else
  \section*{Acknowledgment}
\fi

This work is supported by National Natural Science Foundation of China Program 61876032 and Fundamental Research Program (General Program) of the Shenzhen Science and Technology JCYJ20210324140807019. The authors would like to thank Youngeun Kim for helpful feedback on the manuscript. Yuhang Li completed this work during his prior research assistantship in UESTC.

\ifCLASSOPTIONcaptionsoff
  \newpage
\fi



%

\bibliography{ref,ref_ye}
\bibliographystyle{IEEEtran}

%

\begin{IEEEbiography}[{\includegraphics[width=1in,height=1.25in,clip,keepaspectratio]{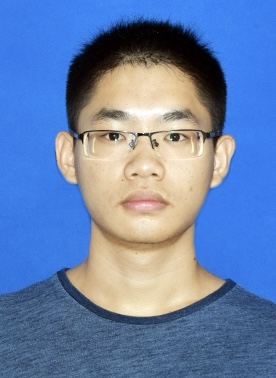}}]{Yuhang Li}
received the B.E. in Department of Computer Science and Technology, University of Electronic Science and Technology of China in 2020. He was a research assistant in National University of Singapore and University of Electronic Science and Technology of China in 2019 and 2021, respectively. Now he is pursuing his Ph.D. degree in Yale University, supervised by Prof. Priyadarshini Panda. His research interests include Efficient Deep Learning, Brain-inspired Computing, Model Compression. 
\end{IEEEbiography}

\begin{IEEEbiography}[{\includegraphics[width=1in,height=1.25in,clip,keepaspectratio]{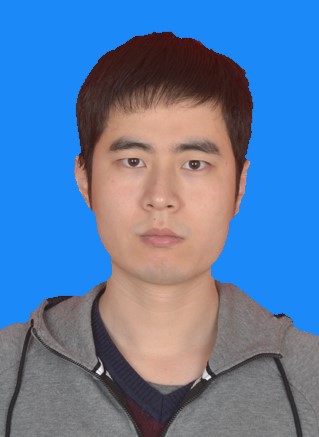}}]{Shikuang Deng}
received his B.E. in the Department of Computer Science and Technology, University of Electronic Science and Technology of China in 2017. He is currently working toward the Ph.D. degree at the University of Electronic Science and Technology of China. His research interests include spiking neural networks, brain control theory, and fMRI data processing.
\end{IEEEbiography}


\begin{IEEEbiography}[{\includegraphics[width=1in,height=1.25in,clip,keepaspectratio]{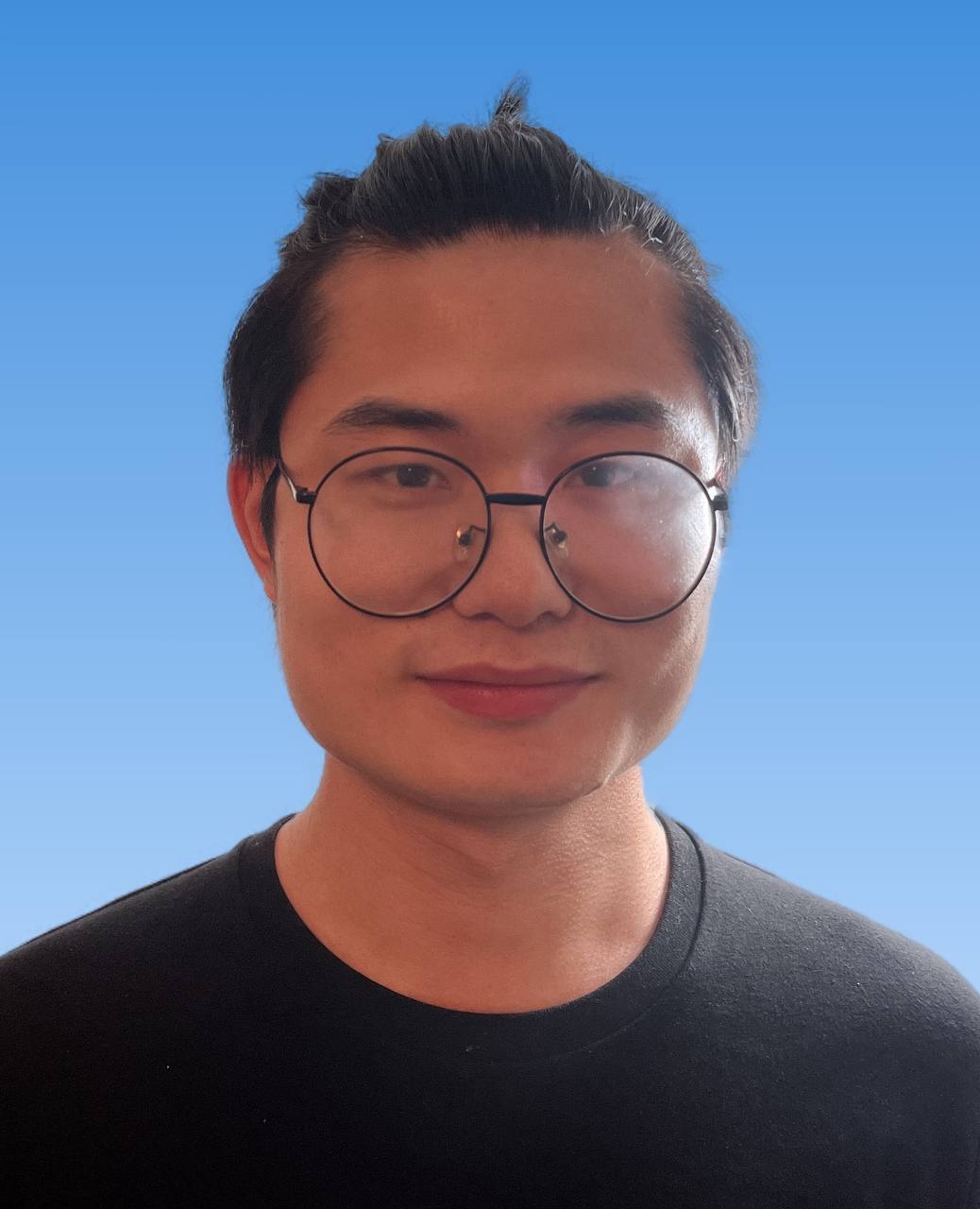}}]{Xin Dong}
is a Ph.D. candidate at Harvard University. His research focuses on efficient and predictable deep learning at the intersection among machine learning, computer architecture and networking. Xin earned bachelor's degree from UESTC in 2017. He was a research assistant at Nanyang Technological University and UC San Diego.
\end{IEEEbiography}

\begin{IEEEbiography}[{\includegraphics[width=1in,height=1.25in,clip,keepaspectratio]{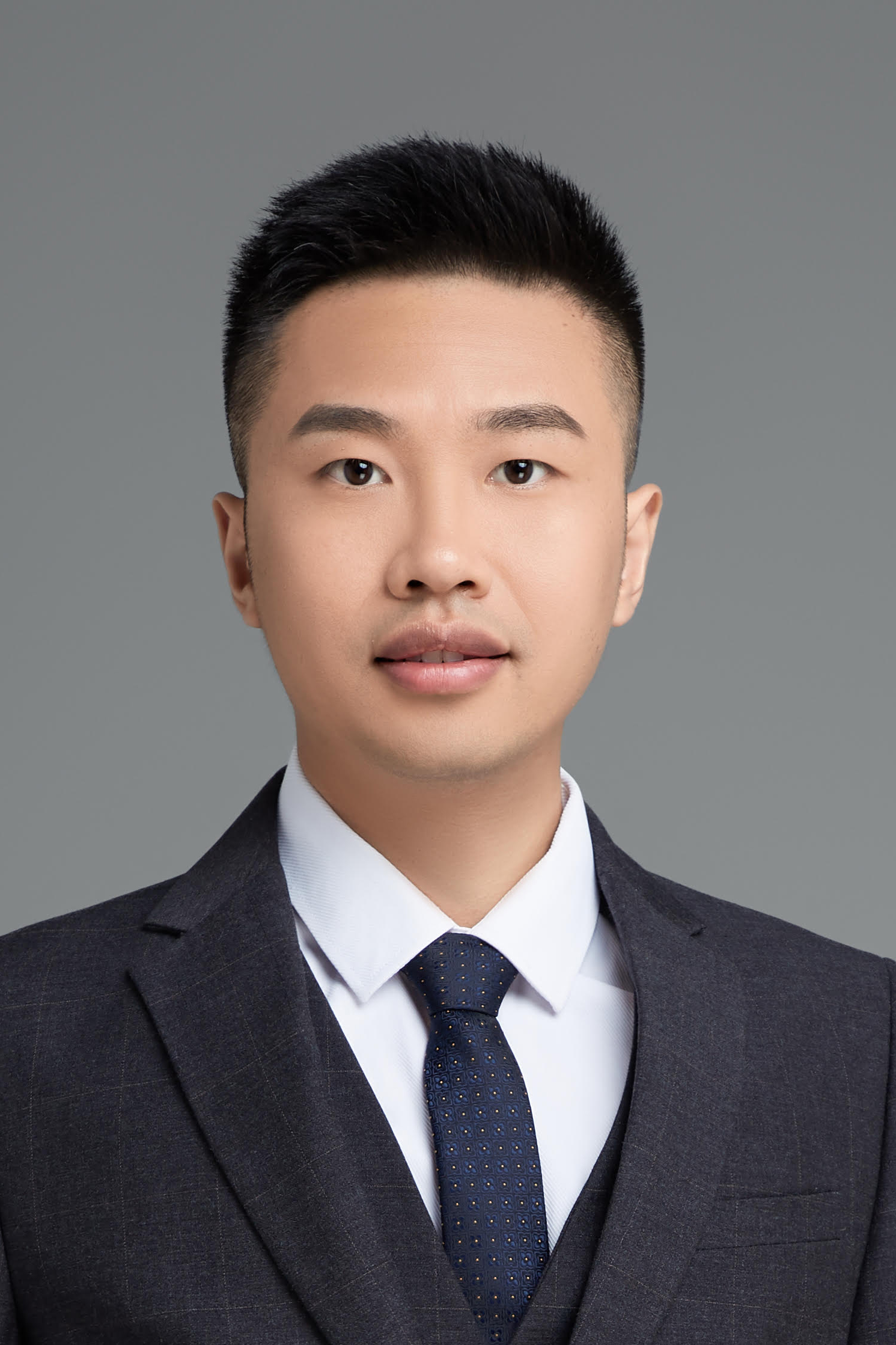}}]{Shi Gu}
received his Ph.D. in the University of Pennsylvania and is a professor in the Department of Computer Science at the University of Electronic Science and Technology of China. His research interests include brain inspired machine learning, network neuroscience and neural encoding models.
\end{IEEEbiography}




\clearpage
\newpage

\appendices
\section{Major Proof}
\label{append_proof}

In order to prove Theorem \ref{theorem}, we need to introduce two lemmas. 

\begin{lemma}
The error term $\mathbf{e}_r^{(b)}$ can be computed using the error from former layer as
\begin{equation}
    \mathbf{e}_r^{(n)} = \mathbf{e}^{(n-1)}\mathbf{BW}^{(n-1)},
    \label{eq_temp1}
\end{equation}
where $\mathbf{B}$ is diagonal matrix whose values are the derivative of ReLU activation. 
\end{lemma}
\begin{proof}
Without loss of generality, we will show that the equation holds for any two consecutive layers $(\ell+1)$ and $(\ell)$. 
Recall that the error term $\mathbf{e}_c^{(\ell+1)}$ is the error between different input activation. Suppose $f(\mathbf{a})=\mathrm{ReLU}(\mathbf{Wa})$ and $\mathbf{e}_r^{(\ell+1)}=f(\mathbf{x}^{(\ell)}) - f(\bar{\mathbf{s}}^{(\ell)})$, we can rewrite $\mathbf{e}_r^{(\ell+1)}$ also using the Taylor expansion, give by
\begin{equation}
    \mathbf{e}_r^{(\ell+1)} =  \mathbf{e}^{(\ell)}\nabla_{\mathbf{x}^{(\ell)}}f + \frac{1}{2}\mathbf{e}^{(\ell),\top} \nabla^2_{\mathbf{x}^{(\ell)}}f\mathbf{e}^{(\ell)} + \mathrm{O}((\mathbf{e}^{(\ell)})^3), 
    \label{eq_local_expand}
\end{equation}
where $\mathbf{e}^{(\ell)}$ is the difference between $\mathbf{x}^{(\ell)}$ and $\bar{\mathbf{s}}^{(\ell)}$.
For the first order derivative, we easily write it as
\begin{equation}
    \nabla_{\mathbf{x}^{(\ell)}}f = \mathbf{BW}^{(\ell)},
\end{equation}
where $\mathbf{B}$ is a diagonal matrix, each element on diagonal is the derivative of ReLU function, which is either 0 or 1. To calculate the second order derivative, we need to differentiate the above equation with respect to $\mathbf{x}^{(\ell)}$. First, matrix $\mathbf{B}$ is not a function of $\mathbf{x}^{(\ell)}$ since it consists of only constants. Second, weight parameters $\mathbf{W}^{(\ell)}$ also is not a function of $\mathbf{x}^{(\ell)}$. Therefore, we can safely ignore the second-order term and other higher order terms in \autoref{eq_local_expand}. Thus, \autoref{eq_temp1} holds.
\end{proof}

\begin{lemma}[\cite{botev2017practical} Eq. (8)] 
The Hessian matrix of activation in layer $(\ell)$ can be recursively computed by
\begin{equation}
\mathbf{H}^{(\ell)} = \mathbf{B}\mathbf{W}^{(\ell)}\mathbf{H}^{(\ell+1)}\mathbf{W}^{(\ell)}\mathbf{B}.
\end{equation}
\end{lemma}

\begin{proof}
Let $\mathbf{H}_{a, b}^{(\ell)}$ be the Hessian matrix (loss $L$ with respect to $\ell$-th layer activation). We can calculate it by 

\newcommand{\FracPartial}[2]{\frac{\partial #1}{\partial #2}}
\newcommand{\fracpartial}[1]{\frac{\partial}{\partial #1}}
\newcommand{\bx}[2]{\mathbf{x}_{#1}^{(#2)}}
\newcommand{\bz}[2]{\mathbf{z}_{#1}^{(#2)}}
\newcommand{\bW}[2]{\mathbf{W}_{#1}^{(#2)}}

\begin{align}
\mathbf{H}_{a, b}^{(\ell)} & = \FracPartial{^2L}{\bx{b}{\ell}\partial\bx{a}{\ell}}=\fracpartial{\bx{b}{\ell}}\left( \sum_i \FracPartial{L}{\bx{i}{\ell+1}}\FracPartial{\bx{i}{\ell+1}}{\bx{a}{\ell}} \right) \\
& = \sum_i \fracpartial{\bx{b}{\ell}} \left( \FracPartial{L}{\bx{i}{\ell+1}} \FracPartial{\bx{i}{\ell+1}}{\bz{i}{\ell}} \FracPartial{\bz{i}{\ell}}{\bx{a}{\ell}}\right) \\
& = \sum_i \bW{i,a}{\ell}  \fracpartial{\bx{b}{\ell}} \left(\FracPartial{L}{\bx{i}{\ell+1}} \FracPartial{\bx{i}{\ell+1}}{\bz{i}{\ell}} \right).
\end{align}
Note that term $\FracPartial{\bx{i}{\ell+1}}{\bz{i}{\ell}}$ is the derivative of ReLU function, which is a constant and can be either 0 or 1, therefore it has no gradient with respect to $\bx{b}{\ell}$. Further differentiating the above equation, we have
\small{
\begin{align}
\mathbf{H}_{a, b}^{(\ell)} & =  \sum_i \bW{i,a}{\ell} \left(  \FracPartial{\bx{i}{\ell+1}}{\bz{i}{\ell}} \FracPartial{^2L}{\bx{b}{\ell}\partial\bx{i}{\ell+1}}\right) \\
& = \sum_i \bW{i,a}{\ell} \left[ \FracPartial{\bx{i}{\ell+1}}{\bz{i}{\ell}} \left( \sum_j \FracPartial{^2L}{\bx{j}{\ell+1}\partial\bx{i}{\ell+1}}\FracPartial{\bx{j}{\ell+1}}{\bx{b}{\ell}}\right) \right] \\
& = \sum_i \bW{i,a}{\ell} \left[ \FracPartial{\bx{i}{\ell+1}}{\bz{i}{\ell}} \left( \sum_j \bW{j,b}{\ell} \FracPartial{^2L}{\bx{j}{\ell+1}\partial\bx{i}{\ell+1}} \FracPartial{\bx{j}{\ell+1}}{\bz{j}{\ell}} \right) \right] \\
& = \sum_{i, j} \bW{i,a}{\ell} \FracPartial{\bx{i}{\ell+1}}{\bz{i}{\ell}} \FracPartial{^2L}{\bx{j}{\ell+1}\partial\bx{i}{\ell+1}} \FracPartial{\bx{j}{\ell+1}}{\bz{j}{\ell}} \bW{j,b}{\ell}.
\end{align}
}
To this end, we can rewrite it using matrix form. Thus, the Lemma holds. 
\end{proof}

Now, we can prove our theorem with above two lemmas. 
\begin{proof}
According to Lemma A.1, the error in the last layer can be rewritten by 
\begin{equation}
    \mathbf{e}^{(n)} = \mathbf{e}^{(n-1)}\mathbf{B}\mathbf{W}^{(n-1)} + \mathbf{e}_c^{(n)}.
\end{equation}
Applying this equation to the second-order objective, we have
\begin{align}
    \mathbf{e}^{(n), \top}\mathbf{H}^{(n)}\mathbf{e}^{(n)} & = \mathbf{e}^{(n-1)}\mathbf{B}\mathbf{W}^{(n-1)} \mathbf{H}^{(n)}\mathbf{W}^{(n-1)}\mathbf{B}\mathbf{e}^{(n-1)} \nonumber\\
    & + 2\mathbf{e}^{(n-1)}\mathbf{B}\mathbf{W}^{(n-1)} \mathbf{H}^{(n)}\mathbf{e}_c^{(n)} \nonumber\\
    & + \mathbf{e}_c^{(n), \top}\mathbf{H}^{(n)}\mathbf{e}_c^{(n)}.
    \label{eq_hessian_decompose}
\end{align}
According to Lemma A.2, the Hessian of activation can be derived recursively as
\begin{equation}
    \mathbf{H}^{(n-1)} = \mathbf{B}\mathbf{W}^{(n-1)}\mathbf{H}^{(n)}\mathbf{W}^{(n-1)}\mathbf{B}.
\end{equation}
Thus, the first term in \autoref{eq_hessian_decompose} can be rewritten to $\mathbf{e}^{(n-1), \top}\mathbf{H}^{(n-1)}\mathbf{e}^{(n-1)}$. 
For the second term, we can upper bound it using Cauchy–Schwarz inequality $(x^\top Ay \le \sqrt{x^\top A xy^\top A y})$ and inequality of arithmetic and geometric mean $(\sqrt{x^\top A xy^\top A y}) \le \frac{1}{2}(x^\top A x+y^\top A y))$. Therefore, the second term is upper bounded by
\begin{align}
    2\mathbf{e}^{(n-1)} & \mathbf{B}\mathbf{W}^{(n-1)} \mathbf{H}^{(n)}\mathbf{e}_c^{(n)} \le \mathbf{e}_c^{(n), \top}\mathbf{H}^{(n)}\mathbf{e}_c^{(n)} \nonumber \\ & + \mathbf{e}^{(n-1)}\mathbf{B}\mathbf{W}^{(n-1)} \mathbf{H}^{(n)}\mathbf{W}^{(n-1)}\mathbf{B}\mathbf{e}^{(n-1)}.
\end{align}
To this end, we can rewrite \autoref{eq_hessian_decompose} as
\begin{align}
    \mathbf{e}^{(n), \top}\mathbf{H}^{(n)}\mathbf{e}^{(n)} & \le 2\mathbf{e}^{(n-1), \top}\mathbf{H}^{(n-1)}\mathbf{e}^{(n-1)} + 2\mathbf{e}_c^{(n), \top}\mathbf{H}^{(n)}\mathbf{e}_c^{(n)} \nonumber\\
    & \le \sum_{\ell=1}^n 2^{n-\ell+1} \mathbf{e}_c^{(\ell), \top}\mathbf{H}^{(\ell)}\mathbf{e}_c^{(\ell)}.
\end{align}
\end{proof}

\section{Experiments Details}

\subsection{ImageNet Pre-training}
The ImageNet dataset~\cite{deng2009imagenet} contains 120M training images and 50k validation images. For training pre-processing, we random crop and resize the training images to 224$\times$224. We additionally apply CollorJitter with brightness=0.2, contrast=0.2, saturation=0.2, and hue=0.1. For test images, they are center-cropped to the same size. For all architectures we tested, the Max Pooling layers are replaced to Average Pooling layers. The ResNet-34 contains a deep-stem layer (i.e., three 3$\times$3 conv. layers to replace the original 7$\times$7 first conv. layer) as described in~\cite{he2019bag}. We use Stochastic Gradients Descent with a momentum of 0.9 as the optimizer. The learning rate is set to 0.1 and followed by a cosine decay schedule~\cite{loshchilov2016sgdr}. Weight decay is set to $10^{-4}$, and the networks are optimized for 120 epochs. We also apply label smooth~\cite{szegedy2016labelsmooth}(factor=0.1) and EMA update with 0.999 decay rate to optimize the model.
For the MobileNet pre-trained model, we download it from \texttt{pytorchcv}\footnote{\url{https://pypi.org/project/pytorchcv/}}.

\subsection{CIFAR Pre-training}
The CIFAR 10 and CIFAR100 dataset~\cite{cifardataset} contains 50k training images and 10k validation images. We set padding to 4 and randomly cropped the training images to 32$\times$32. Other data augmentations include (1)random horizontal flip, (2) Cutout~\cite{devries2017cutout} and (3) AutoAugment~\cite{cubuk2019autoaugment}. For ResNet-20, we follow prior works~\cite{han2020rmp, han2020deep} who modify the official network structures proposed in~\cite{he2016deep} to make a fair comparison. The modified ResNet-20 contains 3 stages with 4x more channels. For VGG-16 without BN layers, we add Dropout with a 0.25 drop rate to regularize the network. For the model with BN layers, we use Stochastic Gradients Descent with a momentum of 0.9 as the optimizer. The learning rate is set to 0.1 and followed by a cosine decay schedule~\cite{loshchilov2016sgdr}. Weight decay is set to $5\times 10^{-4}$ and the networks are optimized for 300 epochs. For networks without BN layers, we set weight decay to $10^{-4}$ and learning rate to 0.005.

\subsection{COCO Pre-training}
COCO dataset~\cite{lin2014microsoft} contains 121,408 images, with 883k objection annotations, and 80 image classes. The median image resolution is 640$\times$480. The backbone ResNet50 is pre-trained on ImageNet dataset. We train RetinaNet and Faster R-CNN with 12 epochs, using SGD with the learning rate of 0.000625, decayed at 8, 11-th epoch by a factor of 10. The momentum is 0.9 and we use linear warmup learning in the first epoch. Batch size per GPU is set to 2 and in total we use 8 GPUs for pre-training. The weight decay is set to $1e-4$. To stabilize the training, we freeze the update in the BN layers (as they has low performance in small batch-size training), and the update in the first stage and stem layer of the backbone.

\subsection{COCO Conversion Details}

For calibration dataset, we use 128 training images taken from the MS COCO dataset for calibration. The image resolution is set to 800 (max size 1333).
Note that in object detection the image resolution is varying image-by-image. Therefore we cannot set a initial membrane potential at a space dimension (which requires fixed dimension). Similar to bias calibration, we can compute the reduced mean value in each channel. Here we set the initial membrane potential as the $T\times \mu(\mathbf{e}^{(i)})$. 
We only convert backbone because the FPN does not have ReLU layers and thus cannot convert to IF, and the head outputs bounding box which requires high precision. Nevertheless, we believe the results in \autoref{tab_detect} can justify the effectiveness of our calibration algorithms. 
Learning hyper-parameters are kept the same with ImageNet experiments. 

\end{document}